\documentclass[manuscript,screen]{acmart} % Remove local acmart.cls from your project to use the official CTAN version
\usepackage{makecell}
\AtBeginDocument{%
  \providecommand\BibTeX{{%
    \normalfont B\kern-0.5em{\scshape i\kern-0.25em b}\kern-0.8em\TeX}}}

\setcopyright{none}              % Removes the ACM copyright statement
\renewcommand\footnotetextcopyrightpermission[1]{} % Removes footnote with conference information
\begin{document}

\title{Survey of Abstract Meaning Representation: Then, Now, Future}

\author{Beherooz Mansouri}
\email{behrooz.mansouri@maine.edu}
\orcid{0000-0002-0400-9761}
\affiliation{%
  \institution{AIIR Lab, University of Southern Maine}
  \city{Portland}
  \state{Maine}
  \country{USA}
}

% \affiliation{%

\renewcommand{\shortauthors}{Behrooz Mansouri}

\begin{abstract}
  This paper presents a survey of Abstract Meaning Representation (AMR), a semantic representation framework that captures the meaning of sentences through a graph-based structure. AMR represents sentences as rooted, directed acyclic graphs, where nodes correspond to concepts and edges denote relationships, effectively encoding the meaning of complex sentences. This survey investigates AMR and its extensions, focusing on AMR capabilities. It then explores the parsing (text-to-AMR) and generation (AMR-to-text) tasks by showing traditional, current, and possible futures approaches. It also reviews various applications of AMR including text generation, text classification, and information extraction and information seeking. By analyzing recent developments and challenges in the field, this survey provides insights into future directions for research and the potential impact of AMR on enhancing machine understanding of human language.
\end{abstract}





\maketitle

\section{Introduction}
Semantic representation refers to the structured modeling of natural language meaning, aiming to capture the underlying semantics of linguistic expressions independently of specific words or syntactic structures. This capability is essential for advancing computational linguistics, natural language processing, and our broader understanding of human language. Various frameworks have been proposed to represent meaning, each developed to capture different aspects of semantics and facilitate diverse downstream applications \cite{abend-rappoport-2017-state, pavlova-etal-2023-structural, sadeddine-etal-2024-survey}.

Central to meaning representation is the concept of events, which serve as the primary units of semantic content. Events are typically expressed through predicates (e.g., verbs) that define the core action or state. These predicates are accompanied by arguments representing participants, properties, or modifiers of the event (e.g., location or time), and are connected by semantic relations. Such structures enable the representation of complex sentence meanings in a concise and formalized manner. As an example, in the sentence ``Joe saw the dog'', \textit{see} is the event, with \textit{Joe} and \textit{dog} as the arguments, representing the subject and the object of the event. 

Meaning representations can be categorized as deep or shallow \cite{sadeddine-etal-2024-survey}. Shallow frameworks, such as Semantic Roles \cite{10.3115/1075218.1075283} and Universal Decompositional Semantics (UDS) \cite{white-etal-2016-universal}, capture the surface-level semantics of text, often focusing on high-level relations between predicates and their arguments. In contrast, deep frameworks go further, representing sub-events and decomposing arguments into finer-grained components. Deep representations are designed to model a wide range of phenomena, including negations, temporal dependencies, causal relationships, comparisons, and modifiers, thus offering a more comprehensive understanding of textual meaning.

Among the deep semantic frameworks, Abstract Meaning Representation (AMR) stands out for its ability to encode the meaning of an entire sentence in a unified, graphical form. Unlike logic-based or distributional approaches, AMR emphasizes semantic abstraction, collapsing syntactic variability into a canonical structure that facilitates high-level reasoning. Originally introduced by Langkilde and Knight (1998) \cite{langkilde-knight-1998-generation} as part of the Nitrogen language generation system, AMR was conceived as a labeled directed graph derived from the PENMAN Sentence Plan Language \cite{mann-hovy-1989-penman}. Early AMRs employed concepts from the SENSUS knowledge base and represented meanings using a simple ``label/concept'' notation. While this early work was primarily explored in the context of language generation, interest in AMR expanded noticeably after Banarescu et al. revisited it for semantic banking (SemBanking) in 2013 \cite{banarescu-etal-2013-abstract}.

In this survey, we present an overview of Abstract Meaning Representation. We begin by introducing AMR and explaining its fundamental concepts and structures. Next, we review approaches for AMR parsing (generating AMRs from text) and AMR-to-text generation (transforming AMRs back into text). We then explore the adaptation of AMR for non-English languages and approaches for multilingual AMR. Finally, we discuss downstream applications of AMR, showing its utility in areas such as text generation, information extraction, information seeking, and text classification.

\section{Abstract Meaning Representation}

Abstract Meaning Representation captures ``who is doing what to whom'' in a sentence. Each sentence is represented as a rooted, directed, acyclic graph with nodes representing concepts and edges showing the relations.\footnote{\url{https://github.com/amrisi/amr-guidelines/blob/master/amr.md}} AMR uses PropBank (Proposition Bank) \cite{10.1162/0891201053630264} frames. PropBank is a layer of annotation added on top of a Treebank that captures the semantic roles of the arguments in a sentence. A Treebank is a corpus of sentences annotated with syntactic structure, typically in the form of parse trees. For example, the Penn Treebank annotates English sentences with part-of-speech tags and phrase structure trees. These trees represent the syntactic organization of sentences, showing how words combine into phrases (like noun phrases and verb phrases) and how these phrases combine into complete sentences. PropBank annotates sentences with predicate-argument structures, indicating which participants in the sentence fill specific roles in relation to a verb (or predicate). These roles are based on the idea of \textit{thematic roles} like agent (the doer of the action), patient (the entity affected by the action), and location (where the action takes place).

PropBank is an important bridge between syntax (as represented in Treebanks) and semantics because it links verb predicates to their roles, enabling understanding of who is doing what to whom in a sentence. Considering the sentence ``The cat sat on the mat'' as an example, the PropBank annotation is \textit{Arg0: The cat (agent)} and \textit{ArgM-LOC: on the mat (location)}. PropBank provides the predicate `sit.01' for the verb `sit' and defines the agent (the cat) and location (on the mat). Abstract Meaning Representation goes further than PropBank by capturing the complete semantics of a sentence in a graph structure. AMR abstracts from syntactic details and focuses entirely on the meaning of the sentence. It not only represents the predicate-argument structure (as PropBank does), but also includes additional semantic information such as negation, modality, quantification, and relations between multiple predicates.

\begin{figure}[t]
    \centering
    \includegraphics[width=\textwidth]{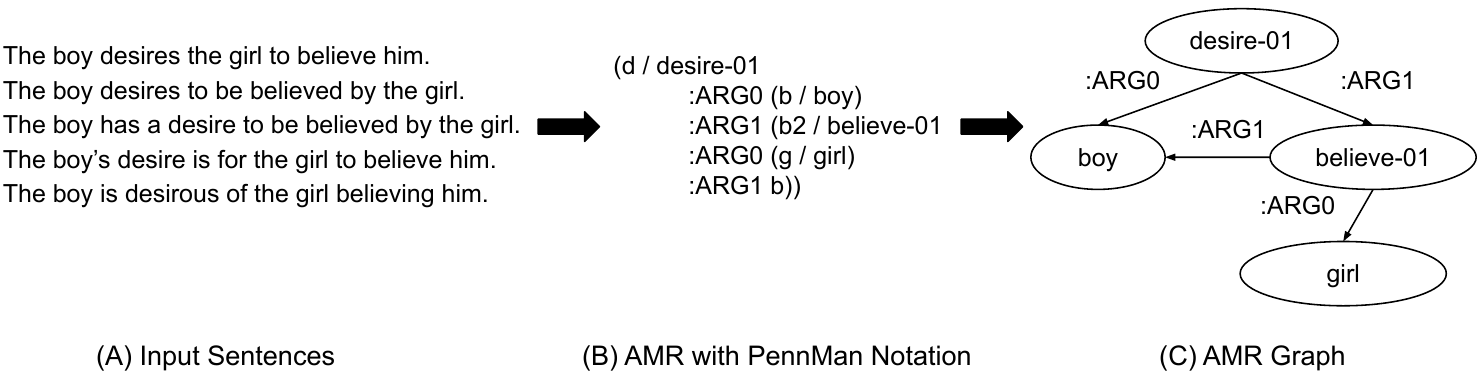}
    \caption{Abstract Meaning Representation for five different input sentences (A) with the same semantics. (B) shows the AMR with PennMan notation and (C) represents the corresponding graph.}
    \label{fig:example1}
\end{figure}
AMR represents sentences as graphs where nodes are concepts (actions, entities, etc.), and edges are relationships between those concepts. AMR also normalizes certain linguistic variations, meaning different syntactic constructions (e.g., active/passive) that convey the same meaning will have the same AMR. Fig. \ref{fig:example1} shows five different sentences that are parsed to a unique AMR.\footnote{Generated using SPRING parser.} In this AMR, `desire-01' has the suffix `-01' representing the specific sense of the verb `desire' based on PropBank; want, crave, desire. The ARG0 connects the verb `desire' to its subject, `boy', and ARG1 connects this verb to `believe', its proposition. The slash (/) in PennMan notation is shorthand for the instance relation. The nodes in this graph show concepts. The concept `b/ boy' refers to instance `boy', and is called `b'. This helps with showing graphs re-entrancy in PennMan notation.

The key features of AMR are:
\begin{enumerate}
    \item \textbf{Graph-Based Representation}: AMR represents sentences as rooted, directed, acyclic graphs (DAGs). The graph nodes correspond to concepts (usually derived from the words in the sentence), and the edges represent the relationships between those concepts. This makes AMR particularly expressive, and capable of representing complex syntactic structures like coordination, subordination, and coreference. Graphical editors such as metAMoRphosED \cite{heinecke-boritchev-2023-metamorphosed} are designed for creating and modifying AMRs.
    \item \textbf{Abstract}: AMR abstracts away from the surface syntactic structure, meaning that sentences with different word orders or syntactic constructions that convey the same meaning will have the same AMR. This abstraction facilitates better generalization for downstream tasks.
    \item \textbf{Uniformity}: AMR aims to provide a uniform, consistent semantic representation for different linguistic constructions. For example, both active and passive voice sentences have the same AMR as long as they convey the same meaning, regardless of differences in syntactic structure.
\end{enumerate}
    
\subsection{What does an AMR Capture?}    
Over time, AMR conventions and standards have been changed to include more linguistic phenomena and increase their application for downstream tasks. Despite these changes, AMR in its foundational format captures a wide variety of semantic and syntactic features, including:
\begin{itemize}
    \item \textbf{Frame Arguments}: AMR uses PropBank framesets to represent the semantic roles of words in a sentence. For example, the frameset ``desire-01'' has slots for the wanter (:arg0), the wanted (:arg1).
    \item \textbf{General Semantic Relations}: In addition to frame arguments, AMR uses nearly 100 general semantic relations to capture relationships between entities, events, and properties. This includes semantic relations such as time, location, and manner.
    \item \textbf{Coreference}: AMR represents coreference by reusing variables to refer to the same entity. In the example from Fig. \ref{fig:example1}, `the boy' and `him' are both represented with variable `b'.
    \item \textbf{Inverse Relations}: AMR uses inverse relations, such as \textit{:arg0-of} and \textit{:quant-of}, to create rooted structures representing the focus of the sentence.
    \item \textbf{Modals and Negation}: AMR represents negation with the \textit{:polarity} relation and expresses modals with concepts.
    \item \textbf{Questions}: AMR uses the concept \textit{amr-unknown} to represent questions and the \textit{:mode} relation to represent yes-no questions, imperatives, and embedded questions.
    \item \textbf{Verbs and Nouns}: AMR uses PropBank framesets to represent both verbs and nouns. 
    \item \textbf{Adjectives and Prepositions}: AMR uses a variety of mechanisms to represent adjectives and prepositions, including PropBank framesets, newly defined framesets, and the \textit{:prep-X} relation for cases where no appropriate relation exists.
    \item \textbf{Named Entities}: AMR represents named entities with the \textit{:name} relation and standardized forms for approximately 80 named-entity types.
    \item \textbf{Copula}: AMR represents copulas using the \textit{:domain} relation.
    \item \textbf{Reification}: AMR allows relations to be treated as first-class concepts through a process called reification.
\end{itemize}

\begin{table*}
    \centering
    \caption{Linguistics aspects that cannot be captured by AMRs.}
    \begin{tabular}{l|l|l}
        \toprule
        \textbf{Aspect}& \textbf{Example Sentence(s)} & \textbf{AMR} \\ \hline
        Verb Tense and Aspect & John eats \& John is eating &\makecell[l]{ (e / eat-01 \\ \hspace{2mm}:arg0 (j / John))}\\\hline
        Syntactic Structure &The boy ate the apple \& The apple was eaten by the boy & \makecell[l]{(e / eat-01\\ \hspace{2mm}:arg0 (b / boy)\\
   \hspace{2mm}:arg1 (a / apple))}       \\ \hline
        Ambiguity&John saw the man with a telescope &\makecell[l]{(s / see-01 \\ \hspace{2mm}:arg0 (j / John) \\ \hspace{2mm}:arg1 (m / man)\\ \hspace{4mm}:manner (t / telescope))}\\\hline
        Word Order&Only John ate the apple \& John only ate the apple &
        \makecell[l]{(e / eat-01 \\ \hspace{2mm}:arg0 (j / John) \\ \hspace{2mm}:arg1 (a / apple))}\\\hline
        Figurative Language&Kick the bucket& \makecell[l]{(k / kick-01 \\\hspace{2mm}:arg0 (p / person) \\\hspace{2mm}:arg1 (b / bucket))}
        \\\hline
        Sentiment and Emotion&I am ecstatic \& I am happy&\makecell[l]{ (h / happy-01\\\hspace{2mm} :arg0 (i / I))}\\\hline
       Intentionality vs. Factuality&He wants to go& \makecell[l]{ (w / want-01\\\hspace{2mm} :ARG0 (h / he)\\\hspace{2mm} :ARG1 (g / go-02\\ \hspace{4mm} :ARG0 h))}\\\hline
       Semantic Variation& He was absent \& He was not present&\makecell[l]{  (a / absent-01\\ \hspace{2mm}:ARG1 (h / he))\\ \\
                    (p / present-02\\ \hspace{2mm}:polarity -\\ \hspace{2mm}:ARG1 (h / he))}\\

        \bottomrule
    \end{tabular}
    
    \label{tab:AMRMiss}
\end{table*}

Although AMR is a strong suit for the mentioned list, there are some aspects that it cannot capture; this includes the following (with examples shown in Table \ref{tab:AMRMiss}):
\begin{itemize}
    \item \textbf{Verb Tense and Aspect:} AMR does not fully capture verb aspect (progressive, habitual, perfect) or nuanced tense distinctions beyond past, present, or future. For example, ``John eats'' and ``John is eating'' are both mapped to the same AMR.
    \item \textbf{Morphological Information:} AMR ignores inflectional morphology such as plural vs. singular forms, verb conjugations, and gender markers. For example, both `cat' and `cats' are represented as `cat'.
    \item \textbf{Syntactic Structure:} AMR abstracts away from syntax, so it does not capture the detailed syntactic structure of sentences. For example, both sentences ``The boy ate the apple'' and ``The apple was eaten by the boy'' will have the same AMR.
    \item \textbf{Ambiguity:} AMR generally represents only one interpretation of an ambiguous sentence, missing other valid meanings. For instance, the sentence ``John saw the man with a telescope'', could mean John used a telescope to see the man, or the man had a telescope. AMR would only capture one interpretation.
    \item \textbf{Word Order:} AMR does not preserve word order, so it misses distinctions that rely on the order of words for meaning or emphasis. ``Only John ate the apple'' vs. ``John only ate the apple'' conveys different meanings, but both would have the same AMR.
    \item \textbf{Figurative Language:} AMR struggles with idioms and metaphors, often either representing them literally or abstracting them away without capturing their full figurative meaning. For example, ``Kick the bucket'' (meaning ``to die'') might be represented literally as a person kicking a bucket.
    \item \textbf{Sentiment and Emotion:} AMR does not handle subtle emotional states or variations in the intensity of emotions well. It can capture basic emotions, but not fine-grained sentiment polarity. Both sentences ``I am ecstatic'' and ``I am happy'' might be represented similarly, though the intensity differs.
    \item \textbf{Intentionality vs. Factuality:} AMRs will not distinguish between the real events that have occurred and intentions. For example, for the sentence ``He wants to go'', in the generated AMR, both verbs `want' and `go' have the same status, where `go' may or may not happen. 
    \item \textbf{Semantic Variation:} AMRs of two different sentences with the same semantic are not identical. For the sentences ``He was absent'' and ``He was not present'' two different AMRs are generated.
\end{itemize}

Although AMR was initially designed only for the English language, later attempts were made to generate AMRs for other languages (see Section \ref{Section_5}). AMRs should also be used carefully for syntax other than natural language. For example, for mathematical formulas, AMRs can be generated for simple equations such as $a=b$, but the AMR notations are not well-defined for more complex formulas \cite{10.1145/3511808.3557567}.

\subsection{AMR Annotations}
The first foundational work that introduced annotation guidelines for AMR was introduced in 2013 \cite{banarescu-etal-2013-abstract}. This work defined the key principles of AMR; how to abstract away from syntax, the use of semantic roles (inspired by PropBank), and the representation of core concepts and their relationships as graphs. AMRs were manually generated for $\sim$5.6K sentences, including the novel \textit{The Little Prince}. With this, the theoretical framework and methodology for creating an AMR corpus was introduced, which became the foundation for the Linguistic Data Consortium (LDC) AMR datasets.

The LDC has released three versions of the AMR annotations, each consisting of thousands of English sentences annotated with AMRs, along with updates and improvements in subsequent releases. These datasets are:
\begin{enumerate}
    \item LDC2014T12\footnote{\url{https://catalog.ldc.upenn.edu/LDC2014T12}} (AMR 1.0) \cite{LDC1}: As a starting point, this dataset contains $\sim$13K English natural language sentences from newswires, weblogs, and web discussion forums. The ontology used in AMR 1.0 was less detailed and comprehensive compared to later versions.
    \item LDC2017T10\footnote{\url{https://catalog.ldc.upenn.edu/LDC2017T10}} (AMR 2.0) \cite{LDC2}: This version provides more precise and consistent guidelines for annotating AMRs, supporting the representation of a wider range of semantic phenomena, such as temporal expressions. It includes $\sim$39K English natural language sentences from broadcast conversations, newswire, weblogs, and web discussion forums.
    \item LDC2020T02\footnote{\url{https://catalog.ldc.upenn.edu/LDC2020T02}} (AMR 3.0) \cite{LDC3}:  This is the most comprehensive, with the largest number of sentences and the most refined and consistent annotations. It includes $\sim$59K English natural language sentences from broadcast conversations, newswires, weblogs, web discussion forums, fiction, and web text.
\end{enumerate}

%Granular AMR Parsing Evaluation Suite (GrAPES) \cite{groschwitz-etal-2023-amr}

\subsection{AMR Enrichment}
While AMR annotations have been updated throughout time, several approaches have been proposed to enrich the AMRs to cover more semantics and meaning. As AMR fails to capture time and aspect, Donatelli et al. \cite{donatelli-etal-2018-annotation} proposed an annotation scheme to support concepts for time such as past, now, future, and concepts for aspects including stable, ongoing, and complete. Later, approaches explored automating the enrichment of AMRs. For example, $A^3$ \cite{ji-etal-2022-automatic}, enriches AMRs to better represent numbers, (in)definite articles, quantification determiners, and intentional arguments. To solve the issue of intentionality versus factuality, a new role \textit{:content} has been added to AMRs \cite{williamson-etal-2021-intensionalizing}, which is interpreted as an intentional operator to represent the scope of modal predicates such as attitude verbs. 
% This process is done by looping through the lemmatized tokens and searching for lemmas that are in the MegaVeridicality dataset \cite{white-etal-2018-lexicosyntactic}.

% For example, for a sentence ``The boy believes a girl is sick,'' the AMR without enrichment is:
% \begin{quote}
%     (b / believe-01\\
%         :ARG0 (b2 / boy)\\
%         :ARG1 (s / sick-05\\
%             :ARG1 (g / girl))
% \end{quote} 
% which faces the issue of invalid inferences given the intentional nature of the attitude verb `believe'. To solve this issue, a similar approach as previous work \cite{williamson-etal-2021-intensionalizing} has been deployed to add a \textit{:content} role, which is interpreted as an intentional operator to represent the scope of modal predicates such as attitude verbs. This process is done by looping through the lemmatized tokens and searching for lemmas that are in the MegaVeridicality dataset \cite{white-etal-2018-lexicosyntactic}.

These graphs are enriched with additional data from the domain to use AMRs for domain-specific applications, mostly in a format similar to AMRs. One approach for enrichment is adding new roles to AMRs. For human-robot interaction, Dialogue-AMR \cite{bonial-etal-2020-dialogue}, developed an inventory of speech acts that represents not only the content of an utterance but the illocutionary force behind it, as well as tense and aspect. Several speech act relations were added to this representation that cannot be directly reflected in AMR. For example, for speech act `Offer', the relation `Offer-SA' is considered, meaning \textit{S} is committed to the feasibility of the plan of action, and \textit{A} is obliged to consider action and respond. Similarly, Gesture AMR \cite{brutti-etal-2022-abstract} uses the idea of Dialogue-AMR for generating enriched AMRs that can capture the meaning of gestures by including gesture acts. 

\begin{figure}[t]
    \centering
    \includegraphics[width=\textwidth]{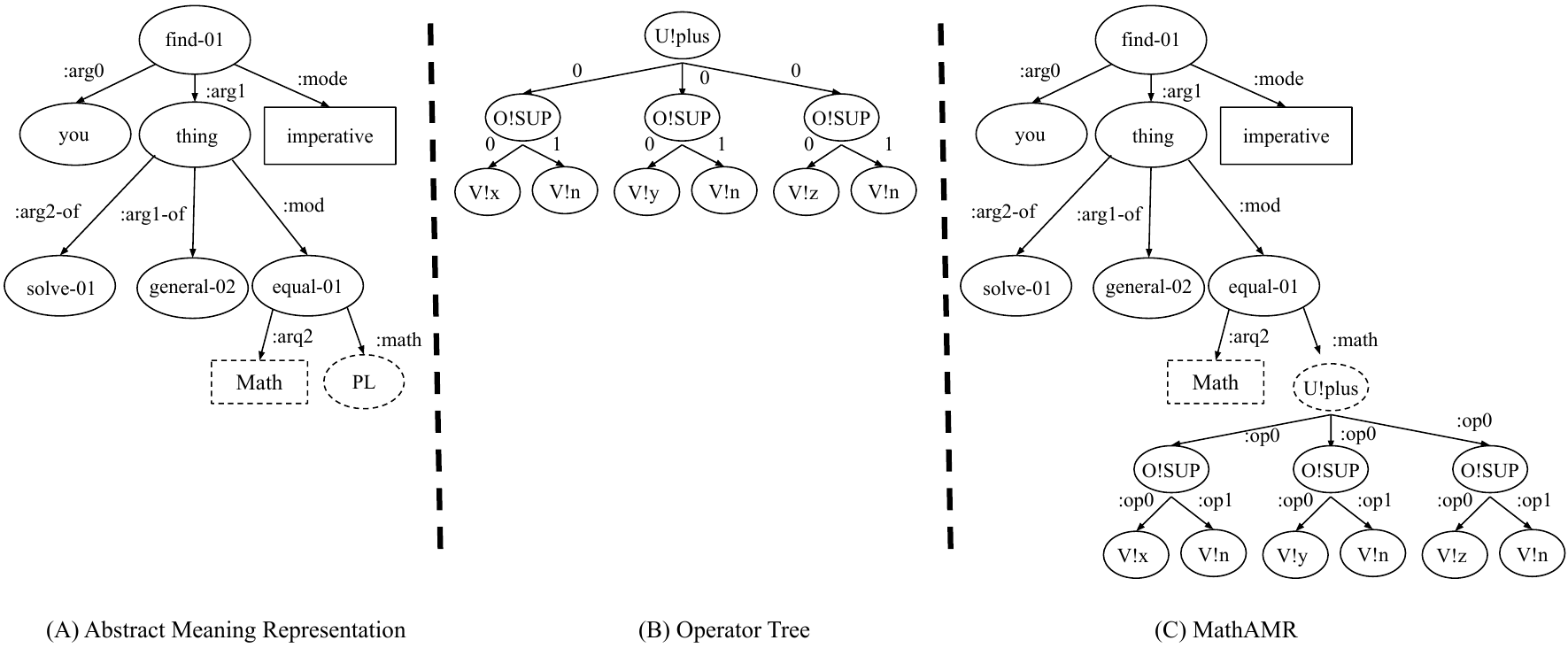}
    \caption{MathAMR \cite{10.1145/3511808.3557567} for input sentence ``Find $x^n+y^n+z^n$ general solution''. (A) AMR is generated for the input text, with the formula replaced with a placeholder (PL) for the formula. Then the operator tree representation of the formula (B), is integrated into the AMR, replacing the placeholder in AMR with the root of the operator tree, resulting in MathAMR shown in (C).}
    \label{fig:mathAMR}
\end{figure}

Another alternative is to integrate other graphs into AMR and have a unified graph representation of input text with text of a specific domain. For instance, the MathAMR \cite{10.1145/3511808.3557567} representation combines the AMR for text representation, and integrates operator trees for formulas in the place of formula to have a unified representation of text and formulas. Operator trees have a similar representation as AMRs, showing what operator should be applied to what operands. The edge labels in operator trees show the order of operands and for non-cumulative operators such as addition, the edge labels for operands are the same (with value 0).  Fig. \ref{fig:mathAMR} shows MathAMR for input text ``Find $x^n+y^n+z^n$ general solution''. First, in (A), the AMR is generated for the modified text, where a placeholder replaces the formula. Then, the operator tree representation of the formula (in (B)), replaces the AMR node representing the formula, shown in (C).

% https://aclanthology.org/2022.law-1.19.pdf
% https://aclanthology.org/2022.lrec-1.169/
% https://aclanthology.org/2024.lrec-main.685/
% https://aclanthology.org/W18-4912/
% https://aclanthology.org/2020.lrec-1.601/
% https://aclanthology.org/2020.dmr-1.1/

% https://arxiv.org/pdf/2404.01129
% https://arxiv.org/pdf/2203.07836
% https://arxiv.org/pdf/2110.05419
% http://nlp.uniroma1.it/spring/
\section{Text-to-AMR (Parsing Task)}
So far, we discussed what AMRs are and how they are annotated. A central challenge, however, is the automatic generation of AMRs for a large volume of natural language text. AMR Parsers are designed to generate AMR graphs for the input natural language text. Words considered semantically light (``do not carry meaning''), such as function words, are typically excluded from the AMR of a sentence. For effective automatic AMR parsing, it is crucial to explicitly represent the alignment between word tokens in the input sentence and the corresponding concepts and relations within its AMR graph. This alignment provides a direct correspondence between the text and its semantic representation.

Early AMR parsing approaches were predominantly alignment-based and employed transition systems to generate AMR graphs. With advancements in neural networks, subsequent approaches, often based on sequence-to-sequence models and graph neural networks, became prevalent. More recently, the application of large language models to AMR parsing has been explored, and future research is likely to investigate their potential more deeply. Before reviewing specific parsing methodologies, it is essential to understand how AMR parsers are evaluated. Therefore, this section will first review the evaluation metrics used for AMR parsers. Then we will review past, current, and possible future approaches for AMR parsing.

\subsection{Evaluation Metrics}
Evaluating the performance of AMR parsers is crucial for assessing their ability to generate AMR graphs from natural language text accurately. One of the earliest reference-based metrics developed for this task is SMatch \cite{cai-knight-2013-SMatch}, which quantifies the similarity between two AMRs by aligning their nodes and counting the number of matching graph triples. These triples are categorized into two types:
\begin{itemize}
    \item $<v_1, rel, val>$, where \textit{val} represents a node value (e.g., (d,instance, chase-01)) 
    \item $<v_1, rel, v_2>$, where both \textit{$v_1$}, \textit{$v_2$} are variables (e.g., (d, ARG0, a))
\end{itemize}
   
While both precision and recall can be calculated using SMatch, most studies report the F1-score as the primary evaluation metric.
To illustrate, Table \ref{tab:SMatch} displays the PENMAN representations and corresponding triples for the gold standard and predicted AMRs of the sentences ``The dog chased the ball'' and ``The canine chased the toy,'' respectively. Precision is computed as the ratio of matched triples to the total number of triples in the predicted AMR (e.g., $\frac{3}{5}=0.6$). Recall is calculated as the ratio of matched triples to the total number of triples in the gold standard AMR (e.g., $\frac{3}{5}=0.6$). The SMatch score is then the harmonic mean of precision and recall, which is 0.6 in this example.
 
\begin{table*}

    \centering
     \caption{AMRs and Triples for a Gold sentence ``The dog chased the ball''. The precision and recall of SMatch are both 0.6, leading to an F1-Score of 0.6.}
    \begin{tabular}{l|l||l|l}
        \toprule
        \multicolumn{2}{c||}{\textbf{AMR}} & \multicolumn{2}{c}{\textbf{Triples}}\\\hline
        \textbf{Gold Standard}& \textbf{Predicted} & \textbf{Gold Standard}& \textbf{Predicted} \\ \hline
        \makecell[l]{(d / chase-01\\\hspace{2mm} :ARG0 (a / dog)\\\hspace{2mm} :ARG1 (b / ball))}&
        \makecell[l]{(d / chase-01\\\hspace{2mm}:ARG0 (a / canine)\\\hspace{2mm}:ARG1 (b / toy))}&
        \makecell[l]{(d, instance, chase-01)\\(a, instance, dog),\\(b, instance, ball)\\(d, ARG0, a) \\(d, ARG1, b)}&
        \makecell[l]{(d, instance, chase-01)\\(a, instance, canine)\\ (b, instance, toy)\\(d, ARG0, a)\\
        (d, ARG1, b)}
        \\
        \bottomrule
    \end{tabular}    
   
    \label{tab:SMatch}
\end{table*}

While SMatch is commonly used in research, it exhibits several limitations. Its reliance on a greedy hill-climbing algorithm for one-to-one node mapping can compromise robustness. Moreover, structural matching alone is insufficient for capturing semantic similarity, and SMatch provides only a single aggregate score, offering little insight into which specific aspects of the parsing may have failed. To address these issues, a fine-grained variant of SMatch was proposed to assess agreement across dimensions such as polarity, semantic roles, and coreference \cite{damonte-etal-2017-incremental}. Subsequently, variants such as S$^2$Match \cite{opitz-etal-2020-amr} and SMatch++ \cite{opitz-2023-SMatch} were introduced. S$^2$Match enables soft semantic matching, relaxing the requirement for exact triple matches, while SMatch++ incorporates three key modifications: 1) Preprocessing steps (e.g., removing duplicate triples), 2) Enhanced alignment through lossless compression, and 3) Scoring mechanisms that consider subgraphs and fine-grained matching. Another related metric, SEMA \cite{anchieta2019semaextendedsemanticevaluation}, uses a breadth-first search approach to compute a maximum score, resulting in a deterministic result and offering a reproducible alternative to SMatch’s greedy approach, with a stricter matching.

While SMatch and its variants are the most widely used metrics to evaluate AMR parsers, a set of alternative metrics has also been proposed. SemBlue \cite{song-gildea-2019-sembleu} is a metric based on the BLEU score that linearizes each AMR using breadth-first traversal, treating nodes as unigrams and connected node-relation pairs as bigrams. Opitz et al. \cite{opitz-etal-2020-amr} conducted a comparative analysis of SemBleu and SMatch based on fundamental evaluation principles, including the requirement that only semantically equivalent graphs should achieve the maximum score, and the principle of symmetry. Their analysis revealed that SemBleu, in contrast to SMatch, violates some of these principles. The Granular AMR Parsing Evaluation Suite (GrAPES) \cite{groschwitz-etal-2023-amr} addresses the limitations of SMatch by providing a more granular evaluation. GrAPES evaluates parsers across a diverse spectrum of linguistic phenomena, such as generalization to unseen labels and structures, and coreference resolution, utilizing a combination of established and newly created sentence-AMR pairs.

Beyond string-based metrics, embedding models have also been explored for AMR parsing evaluation. Opitz et al. \cite{opitz-etal-2021-weisfeiler} utilized the Wasserstein-Weisfeiler Lehman kernel (WWLK) to transform AMR graphs into high-dimensional vector representations. Their approach involves iteratively propagating node embeddings by incorporating contextual information and subsequently employing WWLK to calculate the minimum cost of graph transformation. WWLK, in its original formulation, considers only node embeddings (GloVe embeddings) within AMR graphs, ignoring edge labels. Therefore, WWLK$_\theta$ was extended to incorporate AMR edge labels learning, which requires additional training data. AMRSim \cite{shou-lin-2023-evaluate} is another recent metric based on Graph Neural Networks (GNN) that adopts the pre-trained language model BERT as the backbone and incorporates GraphNN adapters to capture the structural information of AMR graphs.

\begin{figure}[t]
    \centering
    \includegraphics[width=0.8\textwidth]{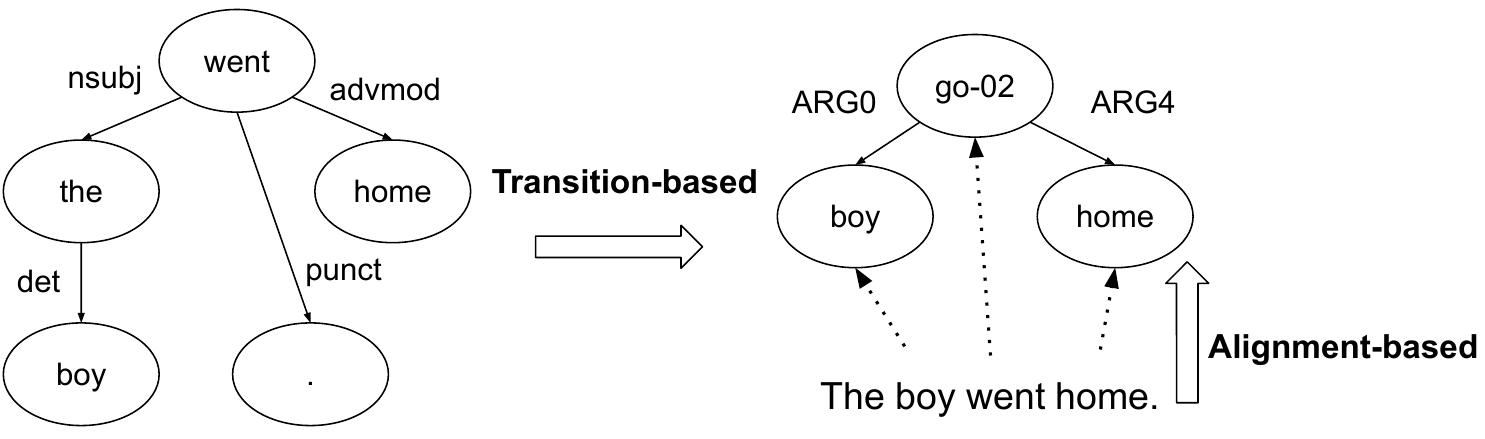}
    \caption{Transition-based and Alignment-based approaches for AMR Parsing.
    Transition-based methods incrementally generate AMRs through a sequence of actions, often guided by a dependency tree (Left). Alignment-based methods first align text spans to AMR nodes and then build the graph based on these alignments (Right).}
    \label{fig:aligntrans}
\end{figure}

\subsection{Then: Alignment and Transition-based Parsing}
In the early days of AMR parsing, alignment-based and transition-based approaches were the predominant strategies. Although these methods were initially proposed as the main strategies for AMR parsing, subsequent models have also been developed from these foundational approaches, including joint alignment–parsing models and neural transition parsers that integrate alignment learning. Alignment-based methods first align spans of the input text to AMR nodes using rule-based heuristics and then perform concept identification and relation extraction based on these alignments.  In contrast, transition-based approaches generate AMRs incrementally by predicting a sequence of actions (e.g., shift, reduce, or arc operations) that build the graph structure, often guided by an intermediate dependency tree. Fig. \ref{fig:aligntrans} illustrates the core ideas behind these approaches.

JAMR \cite{flanigan-etal-2014-discriminative} is widely recognized as the first AMR parser that operates in a two-stage pipeline. First, it maps the concepts in the sentence to concept nodes in the graph using a rule-based semi-Markov model for sequence labeling. Then, it determines the relationships between these concepts by finding the highest-scoring connected subgraph. To find this subgraph, JAMR uses a Maximum Preserving, Simple, Spanning, Connected Subgraph (MSCG) algorithm, which is a novel modification of Kruskal's minimum spanning tree algorithm. This algorithm constructs an initial subgraph that satisfies the preserving, simple, and connected constraints. If the resulting subgraph violates the deterministic constraint, a technique called Lagrangian Relaxation is employed to iteratively adjust the edge scores to enforce this constraint.

Early transition-based parsers initially constructed a dependency tree from the input text and then used a transition system to convert this tree into an AMR graph \cite{wang-etal-2015-transition}. These approaches generate the graph incrementally, resulting in higher graph validity. For instance, AMR-EAGER \cite{damonte-etal-2017-incremental} processes an input sentence from left to right, in a manner similar to transition-based dependency parsers. This parser has three main components: a stack, which holds the nodes of the AMR graph that have been partially constructed; a buffer, which contains the indices of the word tokens from the input sentence that are yet to be processed; and a set of edges, which represents the AMR relationships that have been established so far. The parser starts with an initial configuration and applies a sequence of transition actions (Shift, LArc, RArc, Reduce) until it reaches a terminal configuration, where the entire sentence has been processed and the AMR graph is complete.

CAMR parser \cite{wang-etal-2016-camr} is another transition-based approach to AMR parsing. However, unlike AMR-EAGER, CAMR operates by transforming a dependency parse tree of an input sentence into its corresponding AMR graph. CAMR uses a set of transformation actions performed when processing nodes and edges, as well as actions to infer concepts that do not directly correspond to any word in the sentence.

\subsection{Now: Neural AMR Parsing and BART}
\begin{figure}[t]
    \centering
    \includegraphics[width=0.8\textwidth]{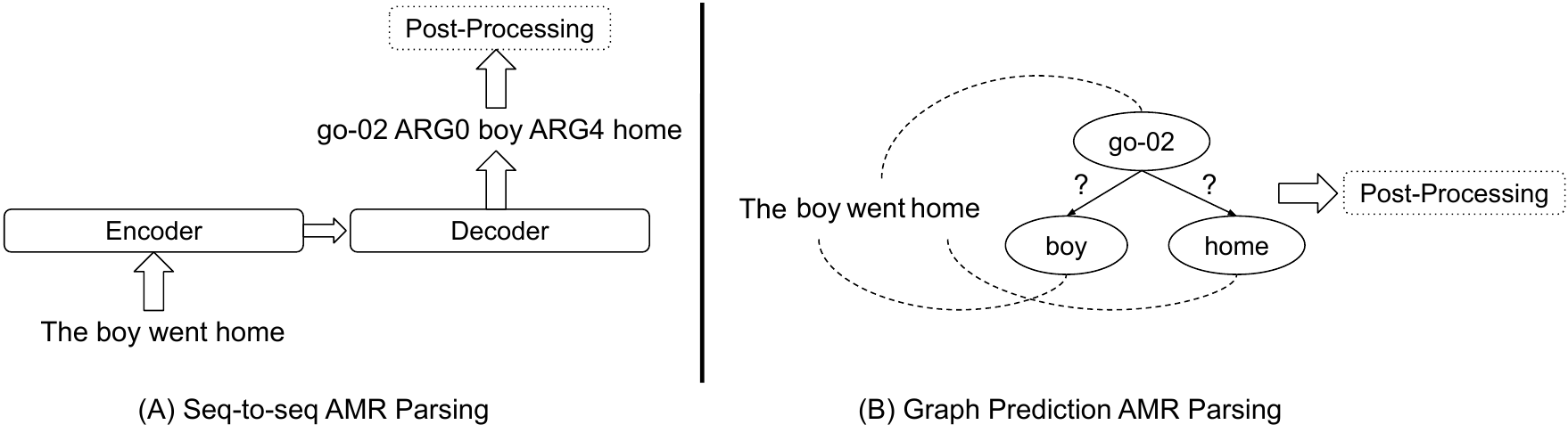}
    \caption{Overview of current AMR parsing approaches. (A) Seq-to-seq models generate a linearized AMR, and (B) Graph prediction directly predicts AMR nodes and edges. Both approaches typically involve pre- and post-processing steps.}
    \label{fig:neuralAMRParsing}
\end{figure}

With the advances in neural network applications, AMR parsing has been explored using these models. Common approaches treat this task either as a sequence-to-sequence problem or as a graph prediction problem.  Fig. \ref{fig:neuralAMRParsing} illustrates these approaches: (A) seq-to-seq models generate a linearized AMR (commonly via depth-first search), and (B) graph prediction models directly predict the AMR graph from the input text.  In both approaches, pre-processing steps serve as a solution for data sparsity, with a set of post-processing steps for AMR validation.

Peng et al. \cite{peng-etal-2017-addressing} were among the first to explore seq-to-seq models for AMR parsing. The architecture consists of an LSTM-based encoder-decoder framework enhanced with an attention mechanism where the encoder processes the input sentence and a decoder generates the linearized AMR. The encoder utilizes a bidirectional RNN, incorporating a backward RNN to capture information from the reverse order of the input sequence. The decoder, at each step, employs an attention mechanism to compute a weighted sum of the input hidden layers, allowing it to focus on relevant parts of the input sequence for generating the output. Data sparsity was addressed by mapping low-frequency concepts and entity subgraphs to a reduced set of category types (e.g., mapping all date entity subgraphs to the category \texttt{DATE}). Although their categorization step was useful in correctly generating AMRs, the overall approach was less effective than previous models such as CAMR.  Similarly, NeuralAMR \cite{konstas-etal-2017-neural} employs a seq-to-seq architecture using a stacked-LSTM with global attention and an unknown word replacement mechanism. This model applies several preprocessing steps, including categorization (using finer categories than Peng et al.) and anonymization of named entities, dates, and quantifiers to reduce data sparsity. One key innovation in NeuralAMR is paired training procedure using a self-training millions of weakly labeled data, and then fine-tuning with human annotated pairs.

As with many other NLP tasks, the introduction of Transformers led to major improvements in AMR parsing, and transformer-based models have become the most widely used. Among transformer-based models, the BART \cite{DBLP:journals/corr/abs-1910-13461} model is commonly used as an encoder-decoder framework. Pre-trained for denoising English text, BART serves as a strong foundation for AMR parsing.  For example, SPRING (Symmetric PaRsIng aNd Generation) \cite{bevilacqua2021one} directly maps an input sentence to its corresponding AMR graph. It applies depth-first search for graph linearization and expands BART’s vocabulary to support AMR tokens; special tokens are used to represent variables in the linearized graph, which helps handle co-referring nodes and prevent data loss. SPRING also incorporates recategorization to reduce vocabulary size. Similarly, the BART architecture was extended in AMRBART \cite{bai-etal-2022-graph} via fine-tuning for both AMR-to-text and text-to-AMR tasks. Unlike SPRING, AMRBART pre-trains the BART model on AMR graphs using two denoising approaches; one at the node/edge level and another at the sub-graph level. For joint AMR graph and text pre-training, four tasks are defined in which the model denoises either the graph or the text using the other modality as context, in both original and noisy input modes.

AMR parsing can also be framed as a graph prediction task. The AMR-GP approach \cite{lyu-titov-2018-amr} jointly predicts both the concept nodes (representing entities and actions) and the relations (edges) between them, while simultaneously handling the alignment between input words and graph nodes. In contrast to traditional methods that treat alignment and parsing as separate steps, this model jointly predicts the entire AMR structure. Similarly, the Sequence-to-Graph Transduction (STOG) model \cite{zhang-etal-2019-amr} view ARM parsing as a two-stage process of predicting nodes and edges. First, node prediction is performed using an Extended Pointer-Generator Network, and second, edge prediction is carried out with a Deep Biaffine Classifier, with the training objective jointly minimizing the loss over nodes and edges. A key characteristic of STOG is that it is aligner-free, unlike many other AMR parsers that rely on pre-trained aligners for word-concept alignment. During pre-processing, AMR graphs are linearized into sequences, often by converting the graph into a tree through the duplication of nodes with reentrant relations and then performing a traversal like depth-first search. Post-processing is then used to recover the full AMR graph from the predicted sequence by restoring variables, wiki-links, and handling co-referring nodes. 

The final category of approaches for AMR parsing are hybrid models. Action-Pointer Transition (APT) parser \cite{zhou-etal-2021-amr} introduces an action-pointer mechanism to leverage the advantages of both transition-based and graph generation-based approaches. The model aligns parsing actions with the input sentence using a cursor-based pointer mechanism. The cursor moves sequentially through the sentence, predicting actions to build the AMR graph incrementally. A pointer network creates edges by referencing previously generated nodes, embedding graph structure directly into the decoding process. A single transformer architecture with self-attention propagates graph information efficiently, modeling both the action generation and the pointer prediction. This approach enables dynamic and interpretable graph construction while maintaining alignment with the input text. Later, the authors introduced StructBART \cite{zhou-etal-2021-structure} that integrates pre-trained BART with a transition-based system to improve AMR parsing. The proposed transition system for AMR parsing is designed to make the most of pre-trained decoders like BART while simplifying the structure compared to previous methods. 

\subsection{Future: Large Language Model Parsers}
Given the success of large language models in various NLP tasks, researchers have begun investigating their ability to handle AMR parsing.  Ettinger et al. \cite{ettinger-etal-2023-expert} examined whether models such as GPT-3, ChatGPT, and GPT-4 can generate AMRs under zero-shot and few-shot settings. In the zero-shot configuration, the prompt was set to ``Provide an AMR (Abstract Meaning Representation) parse for this sentence,'' with a system message (for ChatGPT and GPT-4) instructing, ``You are an expert linguistic annotator.'' In the few-shot setting, five example sentences were provided to guide the models. Evaluation focused on two levels: (1) the overall format, including the highest-level nodes and general semantic accuracy, and (2) the finer accuracy within arguments and modifiers. The results indicate that while LLMs can typically generate valid AMRs (with GPT-4 achieving 100\% validity), they struggle to produce correct corresponding AMRs in the zero-shot setting. However, when provided with few-shot demonstrations, all evaluated LLMs not only output 100\% valid AMRs but also achieve higher accuracy in generating detailed AMR structures.

In addition to training models on (text, AMR) pairs, multitask training approaches are emerging. BiBL (Bidirectional Bayesian Learning) \cite{cheng-etal-2022-bibl} is one such method that employs a single-stage multitask framework, leveraging multiple loss functions (transduction, generation, and reconstruction losses) to simplify training while enhancing performance. Similarly, recent work on incorporating graph structure into learned representations is exemplified by LeakDistill \cite{vasylenko-etal-2023-incorporating}. During training, the model uses word-to-node alignments to build a word-based graph that mirrors the AMR structure. This graph information is then \textit{leaked} to the model to guide learning. To address the absence of this leaked information during inference, LeakDistill employs a self-knowledge distillation technique, transferring the knowledge from a teacher model (trained with the graph information) to a student model that only has access to the input text. LeakDistill achieves Smatch scores of 86.1 and 84.6 on AMR 2.0 and AMR 3.0, respectively, positioning it as a current state-of-the-art AMR parser.

Despite the progress in AMR parsing, the primary focus, along with the development of annotated corpora, has traditionally been at the sentence level. Consequently, the research in AMR parsing, has concentrated on processing individual sentences. Extending the principles of sentence-level AMR to encompass entire documents introduces a set of challenges, including cross-sentential coreferene, computational cost, and lack of consistent standard. To address these challenges, DOCAMR \cite{naseem-etal-2022-docamr} proposes a novel representation that utilizes coref-entity nodes to link coreferent mentions across sentences, while also providing specific rules for handling named entities and pronouns. To tackle the computational cost of evaluating these larger graphs, the paper introduces DOCSMATCH, an optimized version of the Smatch metric that leverages sentence alignments to improve efficiency. Furthermore, DOCAMR includes a coreference subscore to specifically assess the accuracy of cross-sentential links.

\section{AMR-to-Text (Generation Task)}
While parsing text into AMR is one task, generating text from a given AMR has also been widely explored. This task benefits many downstream applications, such as summarization (see Section \ref{app_sec}). It can be defined as generating a sentence that conveys the same meaning as the input AMR graph, essentially, recovering the meaning encoded in the AMR in natural language.

A key challenge is that AMR representations often omit certain linguistic aspects of the original text (see Table \ref{tab:AMRMiss}), therefore, regenerating a full and natural sentence may not capture all the missing details. Models for this task are typically evaluated using machine translation metrics such as BLEU \cite{papineni-etal-2002-bleu}, CHRF++ \cite{popovic-2017-chrf}, and METEOR \cite{banerjee-lavie-2005-meteor}, which assess lexical overlap, n-gram precision, and alignment quality between generated outputs and reference texts.
% While parsing text into AMR is one task, generating text from a given AMR has also been widely explored. This task can benefit many downstream tasks, such as summarization (see Section \ref{app_sec}). The task can be defined as: generating a sentence with the same meaning from an AMR graph}. This can be viewed as recovering the meaning encoded in the AMR as a text. 

% The challenge of this task is that AMR often loses several linguistic aspects of the input text (listed in Table \ref{tab:AMRMiss}), and regeneration may not fully capture these missing elements. The evaluation of models for this task is typically performed using metrics common in machine translation, including BLEU \cite{papineni-etal-2002-bleu}, CHRF++ \cite{popovic-2017-chrf}, and METEOR \cite{banerjee-lavie-2005-meteor}. These metrics assess the generated outputs against reference texts by measuring lexical overlap, n-gram precision, and alignment quality. 

% Early approaches for this task relied on rule-based methods to convert an AMR into its corresponding sentence. Moving neural network era, two common architectures, seq-to-seq, and graph-to-sequence models, have been deployed for this task. While these models continue to be extensively studied, the application of mixed seq/graph-to-seq approaches remains an area for further exploration in the future.

Early approaches for AMR-to-text generation relied on rule-based and statistical methods. In the neural network era, two main architectures, seq-to-seq and graph-to-seq models, have been employed. While these models continue to be extensively studied, the development of hybrid approaches that combine both sequential and graph-based representations remains an area for future exploration.

\subsection{Then: Rule-based Generation}
\begin{figure}[t]
    \centering
    \includegraphics[width=0.9\textwidth]{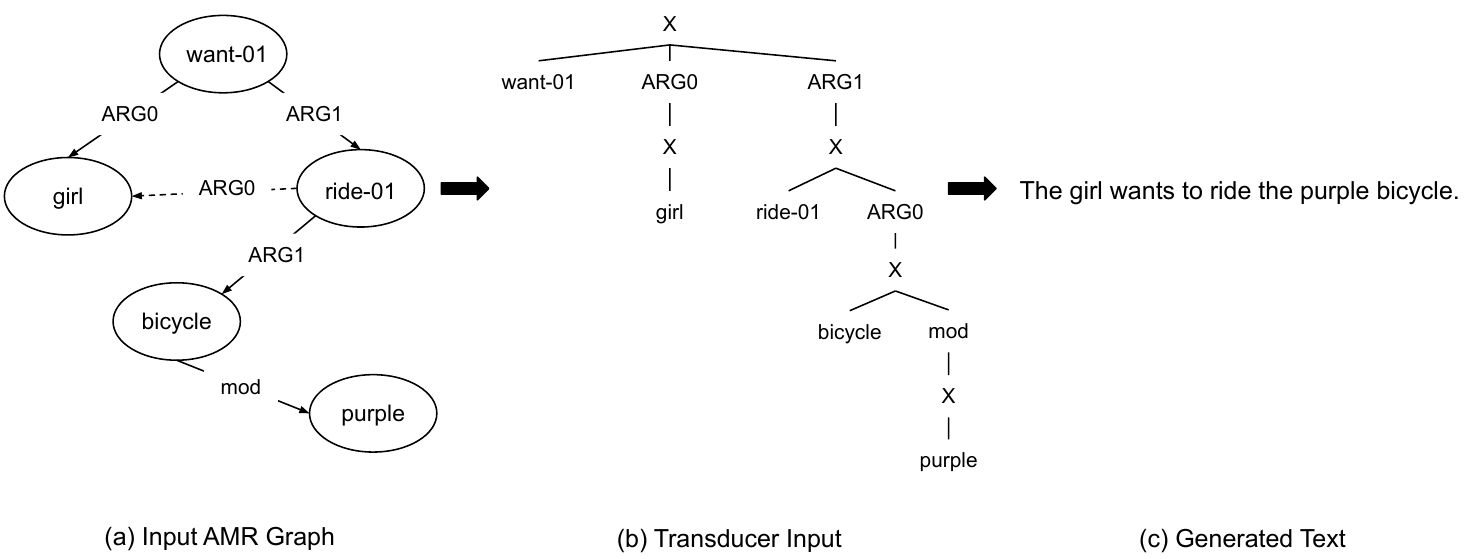}
    \caption{JAMR \cite{flanigan-etal-2016-generation} approach for text generation from Abstract Meaning Representation. (a) The input AMR graph is processed to remove re-entrancies, yielding a tree structure. (b) This tree is then used by a tree-to-text transducer, which (c) generates the final text output.}
    \label{fig:jamrtextgeneration}
\end{figure}

Traditional approaches for AMR-to-text generation have primarily relied on statistical and rule-based methods. The first model proposed for this task was introduced as part of the JAMR system \cite{flanigan-etal-2016-generation}. In JAMR, the AMR graph is first converted into a spanning tree using a breadth-first search traversal. This spanning tree is then processed by a tree-to-string transducer, which generates text based on a set of rules derived from Part-of-Speech (POS) tags and alignment information. Fig. \ref{fig:jamrtextgeneration} provides an overview of this model.
    
Pourdamghani et al. \cite{pourdamghani-etal-2016-generating} proposed an alternative approach that first linearizes the AMR graph into a flat structure and then employs a Phrase-Based Machine Translation (PBMT) system to convert this linearized representation into an English sentence. A key component of this methodology was the learning of a function to linearize the tokens of an AMR graph into an order that resembled English, aiming to reduce the amount of reordering or distortion required during the phrase-based machine translation process. The paper introduced and compared several linearization techniques, with the classifier method demonstrating the most promising results. Their system is trained on AMR-sentence pairs using an alignment algorithm that reduces sparsity by dropping certain structural details (e.g., role edges). This approach has been shown to outperform JAMR on standard AMR datasets, although the tuning of multiple feature functions in PBMT adds engineering complexity. The success of this approach showed the potential of leveraging existing machine translation techniques for the task of AMR-to-text generation, provided that the structural differences between the graph-based AMR and the sequence-based input requirements of PBMT systems could be effectively addressed through a process like linearization.

Building on these methods, Song et al. \cite{song-etal-2017-amr} introduced a heuristic extraction algorithm that learns synchronous node replacement grammar rules from sentence–AMR pairs. At test time, the learned graph-to-string rules are applied via a graph transducer, which collapses the input AMR into an output sentence based on these rules. This method effectively addresses the inherent ambiguity in AMR representations, where a single AMR may correspond to multiple textual realizations, by leveraging synchronous grammar to generate diverse and contextually appropriate sentences.

% Pourdamghani et al. \cite{pourdamghani-etal-2016-generating} proposed an approach that first linearizes the AMR graph into a flat structure, a process known as linearization. They then use a Phrase-Based Machine Translation (PBMT) system to convert this linearized AMR structure into English sentences. The system is trained on AMR-sentence pairs, using an alignment algorithm that reduces AMR sparsity by dropping certain structural details, such as role edges.
% This approach outperforms JAMR on standard AMR datasets. However, the PBMT system, while effective for translation tasks, requires complex tuning of multiple feature functions, which adds engineering complexity to the approach.

% Building on these earlier methods, Song et al. \cite{song-etal-2017-amr} introduced a heuristic extraction algorithm to learn synchronous node replacement grammar rules from sentence-AMR pairs. At test time, the learned graph-to-string rules are applied using a graph transducer. This transducer processes AMRs as input and collapses them into output sentences based on the learned rules, translating the abstract representation into a coherent text. This method addresses the inherent ambiguity in AMR, where a single AMR can correspond to multiple textual realizations. By leveraging synchronous grammar, the approach generates diverse and contextually appropriate sentences from the same AMR input.

\subsection{Now: Neural Generation}
\begin{figure}[t]
    \centering
    \includegraphics[width=0.9\textwidth]{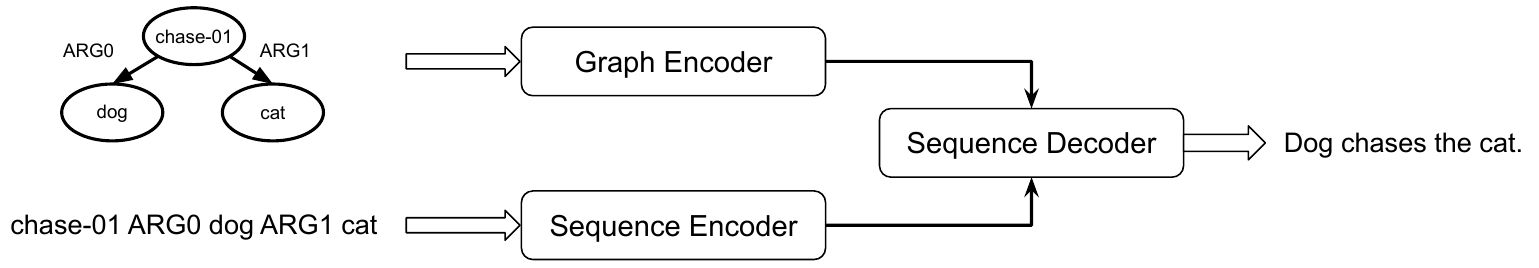}
    \caption{Neural AMR-to-text generation approaches. The top section illustrates graph-to-seq models, where a graph encoder directly processes the AMR to generate text. The bottom section shows seq-to-seq approaches, where a linearized representation of the AMR is processed by a sequence encoder.}
    \label{fig:nerualtextgeneration}
\end{figure}

Similar to AMR parsing, recent approaches for AMR-to-text generation have shifted toward neural architectures, mainly adopting two paradigms: seq-to-seq and graph-to-seq models. Figure \ref{fig:nerualtextgeneration} illustrates these two approaches.

The NeuralAMR model \cite{konstas-etal-2017-neural} was among the first to apply a seq-to-seq framework to the generation task. This model uses a stacked-LSTM encoder–decoder architecture with global attention and an unknown word replacement mechanism. In addition, several preprocessing steps—such as linearization and anonymization of named entities, dates, and quantifiers—are applied to mitigate data sparsity.

Building on this, the Structural Transformer model \cite{zhu-etal-2019-modeling} introduced a structure-aware self-attention mechanism within the transformer framework. As the first transformer-based model for AMR-to-text generation, it leverages linearized AMRs along with explicit edge information to preserve the underlying graph structure. By encoding long-distance relations between concepts via self-attention, it improved performance compared to earlier LSTM-based models. More recently, the SPRING model \cite{bevilacqua2021one} employs BART within a seq-to-seq setup for generation, while models such as BiBL \cite{cheng-etal-2022-bibl} and AMRBART \cite{bai-etal-2022-graph} have unified training for both AMR parsing and text generation tasks, further enhancing their overall utility.

Alternatively, the generation task can be approached as graph-to-seq learning. One of the earliest models in this category is presented by Song et al. \cite{song-etal-2018-graph}, which uses a graph LSTM to directly encode the structure of AMR graphs. This approach bypasses the need for linearization by capturing complex non-local semantic relationships directly from the graph. Expanding on this idea, the GraphTransformer model \cite{wang-etal-2020-amr} adapts the transformer architecture for AMR-to-text generation. In this model, the encoder learns node representations by aggregating information from neighboring nodes through stacked graph attention layers. Each node is assigned two representations: one for aggregating information from outgoing edges (head representation) and another for incoming edges (tail representation). The encoder’s graph attention mechanism effectively preserves edge information without incurring a parameter explosion. To further refine node representations, the model incorporates two graph reconstruction objectives: link prediction for predicting edge labels between nodes and distance prediction between nodes in the graph.

\subsection{Future: Mixed Seq/Graph-to-Seq}
Although current advanced large language models have not been fully investigated for AMR-to-text generation, early work such as GPT-Too \cite{mager-etal-2020-gpt} has explored fine-tuning a GPT-2 model on AMR data. Unlike traditional AMR-to-text approaches that often rely on specialized graph-to-seq models, GPT-Too adapts a general-purpose language model through targeted training adjustments, including cycle-consistency-based re-scoring, to better align the generated text with the underlying AMR structure, thereby improving both semantic accuracy and fluency.

Recent research indicates that while seq-to-seq models with AMR linearization can produce strong results, there is still significant potential in exploring different linearization strategies. For example, Hoyle et al. \cite{hoyle-etal-2021-promoting} found that models trained on fixed canonical linearizations struggled to generalize; introducing random traversal-based linearizations and scaffolding losses improved graph sensitivity, particularly in low-resource settings. Building on these findings, StructAdapt \cite{ribeiro-etal-2021-structural} proposed a novel adapter architecture that integrates graph connectivity into the encoding process. Furthermore, Montella et al. \cite{montella-etal-2023-investigating} explored incorporating relative positional embeddings in transformer models which enhanced generation even when the AMR structure was partially erroneous or missing, suggesting that further exploration of mixed seq/graph representations is necessary.

Previous research has shown that graph-based representations of AMRs generally outperform tree-based (ignoring reentrancies) and linearized representations \cite{damonte-cohen-2019-structural}. However, each representation offers unique advantages. Recent work, such as the DualGen model \cite{hongtwo}, exemplifies this by employing a dual-encoder design, one encoder processes the AMR graph using a Graph Neural Network (GNN) to capture relationships between concepts, while the other processes a linearized sequence of AMR nodes to capture ordering information. DualGen achieves BLEU scores of 51.6 on AMR 2.0 and 51.8 on AMR 3.0, positioning it as the current state-of-the-art for AMR-to-text generation. Combining these representations effectively remains a promising future direction for text generation tasks.

\section{AMR for Non-English Languages} \label{Section_5}
Although AMR was originally developed for English, its design abstracts away from language-specific features such as word order, morphology, and function words, thereby reducing many sources of cross-linguistic variation \cite{banarescu-etal-2013-abstract}. While AMR was not intended to serve as an interlingua, this abstraction has inspired research on its utility as a cross-lingual semantic representation \cite{wein-schneider-2021-classifying}.  Early research compared manually generated AMRs for various languages to their English counterparts, identifying common semantic structures. In practice, AMR corpora for other languages are typically created in one of two approaches. One approach adapts the English annotation guidelines for the target language, with experts annotating data accordingly. In another approach, source-language sentences may be translated into English, and then existing English AMR parsers are applied to generate the AMR graphs.

A related task is \textit{cross-lingual AMR parsing}, where a sentence in any language is converted into its corresponding English AMR graph (with nodes representing English words, PropBank framesets, or special AMR keywords). Damonte and Cohen \cite{damonte-cohen-2018-cross} explored two strategies for this task in Italian, Spanish, German, and Chinese: annotation projection, which leverages word alignments to transfer AMR structure from one language to another, and machine translation, which first translates the input into English before applying an English AMR parser.

\begin{figure}[t]
    \centering
    \includegraphics[width=0.9\textwidth]{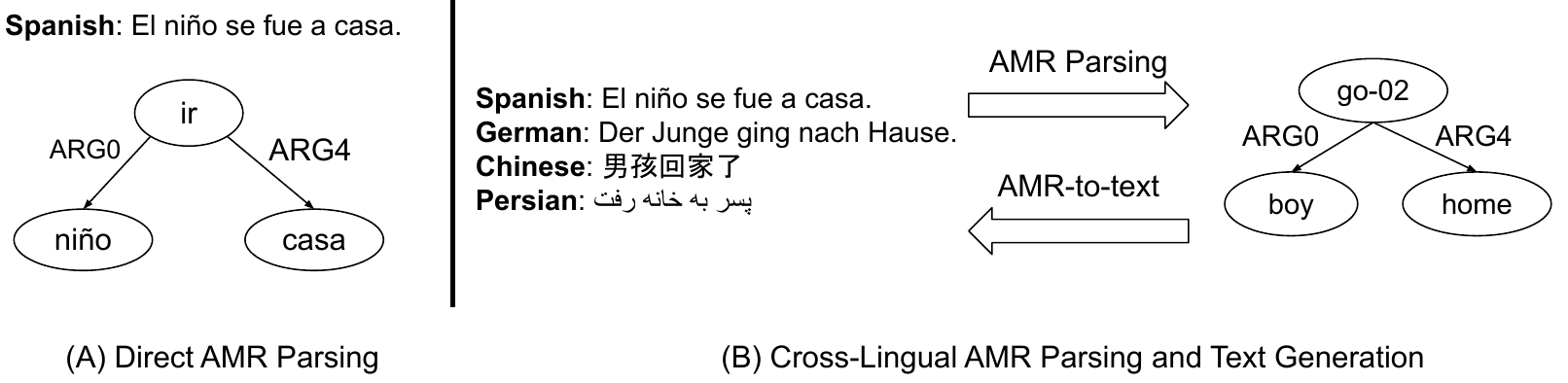}
    \caption{AMR Parsing and AMR-to-text Generation Tasks for Non-English Languages. Approaches shown are: (A) direct AMR parsing using target-language annotation guidelines;  (B) cross-lingual AMR parsing to generate English AMRs from non-English sentences, and AMR-to-text generation to produce sentences in the target language from English AMRs.}
    \label{fig:multilingual}
\end{figure}

Figure \ref{fig:multilingual} illustrates these common approaches. Some methods focus on language-specific annotation and parser development (e.g., training an AMR parser directly on Spanish sentences), while others concentrate on cross-lingual tasks, such as generating English AMRs from non-English text or producing target-language sentences from English AMRs.

\subsection{Then: Exploring Feasibility} 
One of the first studies to explore the compatibility of non-English AMRs with English AMRs examined 100 Chinese and Czech sentences and their corresponding AMRs against English \cite{xue2014not}. The analysis revealed that it is not always feasible to structurally align English AMRs with those from Czech and Chinese; however, refined annotation guidelines can help address some of these discrepancies. Additionally, the compatibility between English and Chinese AMRs was found to be greater than that between English and Czech AMRs. This idea was later extended by developing an AMR parser capable of handling multiple languages \cite{vanderwende-etal-2015-amr}. However, due to the lack of annotated AMR corpora and language-specific specifications, the parser would introduce errors that did not comply with the AMR format. 

Building on these early efforts, researchers turned to leveraging literary texts to construct non-English AMR corpora. For instance, since one of the earliest AMR corpora was based on the novel \textit{The Little Prince}, subsequent work utilized this novel for non-English annotation.  The annotation of the Chinese translation of \textit{The Little Prince} marked the first attempt at building a non-English AMR corpus \cite{li-etal-2016-annotating}.\footnote{While this corpus is also referred to as `CAMR', we use `CHAMR' to avoid confusion with CAMR parser.} This CHAMR corpus comprises 1,562 sentences annotated following English AMR guidelines that were modified to better accommodate Chinese. In CHAMR, argument labels and predicate senses are based on the Chinese Proposition Bank, while other aspects of the annotation follow the English design. For example, if a city's name is written in Chinese, the entity is labeled as  `city'  and is connected via a `name' edge to its Chinese name. 

Later studies further explored the feasibility of AMR annotation for additional languages. In one study, 100 sentences were manually annotated for Turkish by a non-Turkish linguist, who aligned the original English sentences with their literary Turkish translations to create corresponding AMR graphs \cite{azin-eryigit-2019-towards}. In this Turkish corpus, a few sentences share exactly the same AMR structure as their English counterparts; however, most divergences arise from differences in word choice during translation. Turkish appears to be more expressive due to its use of suffixes, which add nuances such as possession markers and intensifiers, and its frequent use of light verbs and multi-word expressions. In English AMR annotation, light verb constructions are typically removed and OntoNotes predicate frames are used to handle verb-particle combinations. Due to Turkish’s highly productive morphology and idiosyncratic features, extra care is required when dealing with multi-word expressions and light verb constructions in AMR annotation.

% A similar study on ``The Little Prince'' has been conducted for the Turkish language \cite{azin-eryigit-2019-towards}, where 100 sentences were manually annotated by a non-Turkish linguist. This linguist aligned the English sentences with their literary translations in Turkish and created the corresponding AMR graphs. A few sentences exhibited identical AMR structures to their English counterparts. However, divergences in the Turkish AMR annotations were primarily due to different word choices in the translation, with Turkish often being more expressive—suffixes add nuances to words, such as possession markers and intensifiers. Unlike English, Turkish contains many light verbs and multi-word expressions. In English AMR, light verb constructions are typically simplified, and OntoNotes predicate frames are used to handle verb-particle combinations. However, given the highly productive nature of Turkish and its unique features, greater caution is needed when dealing with multi-word expressions and light verb constructions in Turkish AMR.

\subsection{Now: Annotation Guidelines and Parsers}
Efforts to develop annotation guidelines for AMR have extended to several non-English languages \cite{migueles-abraira-etal-2018-annotating, azin-eryigit-2019-towards, choe-etal-2020-building, heinecke-shimorina-2022-multilingual}. The primary approach involves adapting the English AMR guidelines to the target language and identifying areas where divergences occur. In many cases, proposed solutions address these discrepancies, although some issues remain open for future research. For example, Spanish AMR annotation has been studied \cite{migueles-abraira-etal-2018-annotating}, focusing on the need for language-specific adjustments, such as handling noun phrase ellipses, third-person possessives and clitic pronouns, the use of “se,” gender distinctions, specific verbal constructions (e.g., verbal periphrases and locative expressions), and double negatives. Another example is the AMR 2.0 - Four Translation Corpus, which comprises Spanish, German, Italian, and Chinese Mandarin translations of a subset of sentences from AMR Annotation Release 2.0 \cite{AB2/5OU0AQ_2022}.

Recent advancements in multilingual AMR parsing and AMR-to-text generation have moved beyond language-specific manual annotations. Modern models leverage neural architectures and multilingual pre-trained models to overcome some limitations of earlier approaches. Instead of relying solely on manual annotation guidelines or alignment strategies between English AMR and non-English corpora, these models employ techniques such as knowledge distillation and multilingual transformer architectures to enable cross-lingual parsing.

For instance, while Damonte and Cohen \cite{damonte-cohen-2018-cross} utilized alignment for cross-lingual AMR parsing, XL-AMR \cite{blloshmi-etal-2020-xl} enables AMR parsing across different languages by leveraging a multilingual transformer model without requiring explicit alignment. XL-AMR incorporates two jointly trained modules for concept and relation identification. The concept identification module detects English words, AMR nodes, and PropBank framesets using a seq-to-seq architecture for improved cross-lingual performance, while the relation identification module reconstructs the AMR graph using a deep biaffine classifier. The model is trained using two strategies: (1) employing silver training data through annotation projection (where English AMR graphs are projected onto target language sentences), and (2) translating English (sentence, AMR) pairs to create gold-standard training data in the target language. The latter approach has yielded better results, consistent with later findings \cite{uhrig-etal-2021-translate}.

XAMR \cite{cai-etal-2021-multilingual-amr} proposed a multilingual AMR parser designed to work across all languages without requiring explicit word-to-node alignments. This approach uses an English parser as a teacher and applies knowledge distillation (both token-level and sequence-level) to train a student model that operates across multiple languages. Due to the lack of human-annotated AMR datasets for non-English languages, the authors generated silver training data by translating English sentences into other languages and applying back-translation consistency checks to ensure quality. The parser underwent multiple pre-training and fine-tuning stageswhile incorporating noisy knowledge distillation techniques.

Multilingual AMR-to-text generation has also been explored. In this task, the goal is to generate text in languages other than English from an AMR.  Fan and Gardent \cite{fan-gardent-2020-multilingual} trained a multilingual AMR-to-text generation model for 21 different languages using EUROPARL multilingual corpus. They generated English AMRs using JAMR, which were then used as inputs for the generation task; the graph encoder and language models were pre-trained on between 400K and 8.2M (graph, text) pairs, depending on the target language. XLPT-AMR \cite{xu-etal-2021-xlpt} represents a zero-shot multilingual AMR parser and generator pre-trained with multi-task learning that includes parsing, generation, and translation tasks. After pre-training, several fine-tuning strategies were explored. The most effective was a teacher-student-based multi-task learning approach, in which stronger English tasks served as `teachers' to guide weaker tasks in German as `students.' Knowledge distillation was applied to align token generation probability distributions between the teacher and student, reducing noise from machine-translated German datasets.

\subsection{Future: Expanding AMR Corpora and Advancing Multilingual Parsers}
Efforts to develop AMR corpora for non-English languages are ongoing. For example, DeAMR \cite{otto-etal-2024-corpus} is an annotated corpus for German created by adapting the English AMR guidelines, while similar initiatives are underway for low-resource languages such as Persian \cite{10.1145/3638288} (PAMR), which comprises 1,020 Persian sentences and their corresponding AMRs. The study highlights challenges in AMR parsing for Persian, including issues related to translation, sense numbers, English tags and expressions, and the unique structure of Persian writing. In addition, MASSIVE-AMR \cite{regan2024massivemultilingualabstractmeaning} represents a major recent contribution, serving as the largest corpus for multilingual AMR parsing to date; it includes 84,000 text-to-graph annotations derived from 1,685 question-answer utterances across 52 languages, based on manual translations from the MASSIVE dataset \cite{fitzgerald-etal-2023-massive}.

The assessment of AMRs in other languages is also gaining attention. Wein and Schneider \cite{wein2024assessing} investigate the applicability of AMRs in cross-lingual contexts, revealing that the source language significantly influences AMR structures. Translation divergences and annotator choices contribute to cross-linguistic differences, indicating that cross-lingual evaluations must account for the source language’s influence to ensure accurate semantic interpretations. Despite challenges posed by lexicalization differences, AMR shows promise for applications in non-English languages, including machine translation, summarization, and event extraction, provided that these cross-linguistic nuances are carefully addressed.

AMR parsing and generation for non-English languages, especially low-resource ones, remain active research areas with new approaches emerging. Meta-XMAR \cite{kang-etal-2024-cross-lingual} introduces meta-learning and joint-learning strategies for cross-lingual AMR parsing, focusing on underrepresented languages such as French, Chinese, Korean, Persian, and Croatian. This approach employs a seq-to-seq framework using the mBART model \cite{tang2020multilingualtranslationextensiblemultilingual} that linearizes AMR graphs by removing variables and wiki links. By simulating k-shot learning during training, the parser adapts quickly to new languages, iteratively updating model parameters to generalize across languages.

Martinez and Parmentier \cite{soto-martinez-etal-2024-generating-amrs} propose two efficient techniques to improve AMR-to-text generation for both high-resource and low-resource languages. They introduce Hierarchical QLoRA, a variation of curriculum learning that iteratively refines a multilingual model into monolingual models using Low-Rank Adaptation (LoRA) and 4-bit quantization for memory efficiency. Their approach fine-tunes a multilingual mT5-large model in a tree-like structure across four levels, progressively training on smaller subsets of languages to balance cross-lingual transfer and regularization. Two grouping strategies are explored: one based on maximizing the distance between languages to serve as a regularizer and the other based on phylogenetic relationships to promote transfer between closely related languages.

Overall, future trends in non-English AMR research are likely to focus on expanding large and diverse multilingual AMR corpora, especially for low-resource languages, and on refining cross-lingual parsing and generation techniques through advanced neural architectures. Emphasis will likely shift toward leveraging multilingual pre-trained models, zero-shot learning, and meta-learning to overcome resource limitations and improve semantic consistency across languages.

\section{AMR Applications} \label{app_sec}
AMRs are used in a wide range of natural language processing tasks. These tasks can be broadly categorized into text-to-text generation, text classification, information extraction, information seeking, and more. This section explores several applications of AMRs across these domains.

\subsection{Text-to-Text Generation}
Several natural language processing tasks involve transforming input text into output text, such as summarization and machine translation. AMR has been explored as a valuable tool for these tasks, offering a structured semantic representation of text.

\begin{figure}[t]
    \centering
    \includegraphics[width=\textwidth]{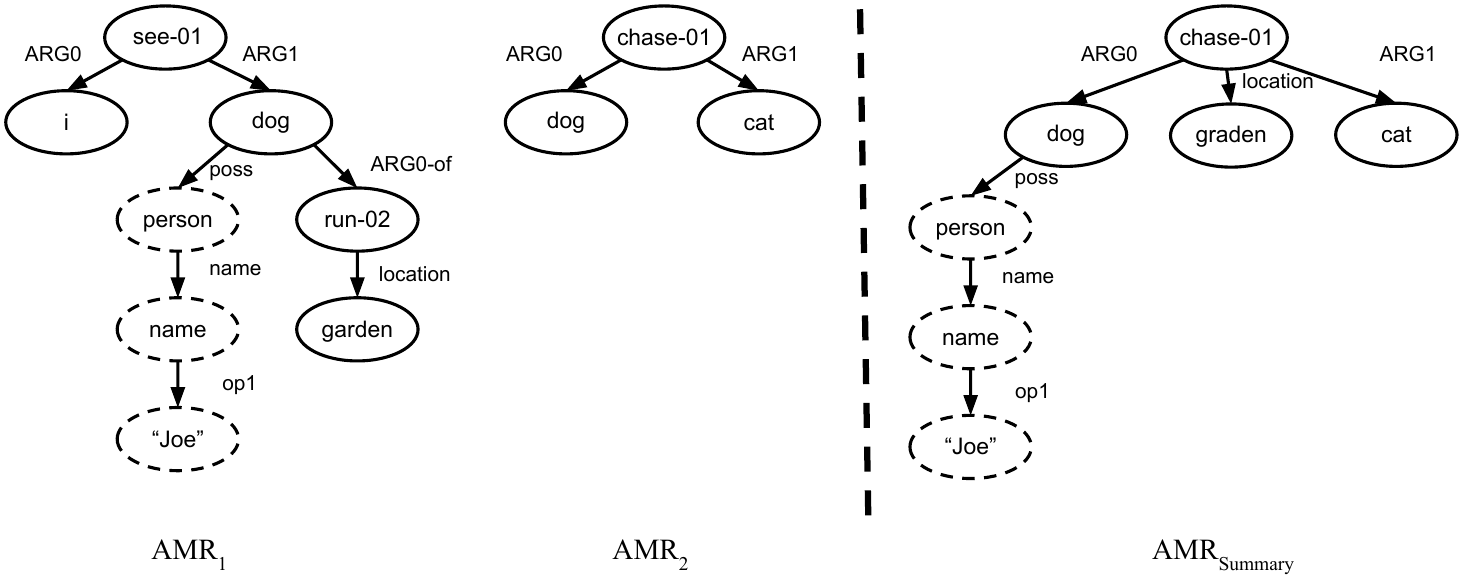}
    \caption{Application of AMR for summarization (Adapted from Liu et al \cite{liu-etal-2015-toward}). $AMR_1$ shows the AMR for the sentence ``I saw Joe’s dog, which was running in the garden.'' and $AMR_2$ shows the AMR for the sentence ``The dog was chasing a cat.'' The summary graph is represented as $AMR_{Summary}$.}
    \label{fig:SummaryAMR}
\end{figure}

\textbf{Summarization.} The application of AMRs for summarization began with abstractive summarization, where the goal is to generate summaries by rephrasing the main ideas of the input text into new sentences. Liu et al. \cite{liu-etal-2015-toward} introduced an approach in which each sentence is first parsed into an AMR graph. These graphs are then merged and transformed into a single summary AMR graph that is passed to a text generator to produce the final summary. An example of this process is shown in Fig. \ref{fig:SummaryAMR}, which summarizes two sentences: ``I saw Joe’s dog, which was running in the garden'' and ``The dog was chasing a cat.'' In the merging step, graph fragments are collapsed into unified concepts by merging nodes with the same label across different sentences. The sub-graph prediction is formulated as an integer linear programming optimization problem to select the most important parts of the merged AMR graph for final summary generation. A similar approach is used for multi-document summarization \cite{liao-etal-2018-abstract}, where the summary is generated from the PENMAN representation of the summary AMR.

Despite the progress made, early work on summarization faced challenges in generating fluent language from AMRs, often producing outputs that resembled a bag-of-words collection. To address these issues, the AMR2Text model \cite{hardy-vlachos-2018-guided} introduced a guided language generation approach. Using a seq-to-seq model with attention, the guided model combined decoder probability distributions with those derived from the summary AMR. Sentences exhibiting the highest similarity to the summary AMR graph were identified using the Longest Common Subsequence (LCS) method, and n-gram probabilities were interpolated to refine word selection. Experimental results demonstrated that this guided approach produced more coherent summaries compared to unguided baselines.

AMRs have also been explored for extractive summarization, which aims to retain all critical information by directly selecting salient portions of the text while avoiding incoherencies or unresolved anaphoric references. Mishra and Gayen \cite{MISHRA2018178} proposed a method that applies pairwise coreference resolution followed by AMR generation for individual sentences. The resulting AMRs are merged into a comprehensive graph representing the entire text, from which key summary segments are then extracted.
% Evaluations on the CNN/DailyMail dataset demonstrate the effectiveness of this approach in producing coherent and lossless summaries, showing its potential for condensing information in a way that preserves the original content's integrity. 

AMRs are particularly useful for handling complex entity interactions, such as in TV series transcripts. AMRTVSumm \cite{hua-etal-2022-amrtvsumm} leverages AMRs to represent individual scenes within an episode. These scene-level AMRs are subsequently integrated into a hierarchical encoder-decoder model that incorporates a novel cross-level cross-attention mechanism. This approach builds on dialogue-specific AMRs introduced by Bai et al. \cite{bai-etal-2021-semantic}, which enhance traditional AMRs by including speaker nodes, utterance nodes, and improved coreference handling to better capture conversational dynamics.

% When applied to the SummScreen datasets, AMRTVSumm consistently outperforms a hierarchical transformer baseline but falls short of the state-of-the-art DialogLM, which benefits from extensive in-domain pretraining. The study highlights the potential of AMR in enhancing the summarization of complex dialogues but notes the importance of combining AMR with robust, dialogue-specific pretrained models.

\textbf{Machine Translation.} Defined as the task of converting source text into a target language, machine translation has also benefited from AMR-based approaches that integrate semantic information from AMR graphs into translation models. Early work by Song et al. \cite{song-etal-2019-semantic} used a graph recurrent network to encode AMR graphs, alongside a Bi-LSTM to encode the source text. The decoder (a doubly attentive LSTM) generated translations by attending to both graph-based and sequential encodings. This dual-encoding strategy effectively leveraged the semantic structure provided by AMRs, laying the groundwork for future innovations. A similar approach was later proposed by modifying the model architecture and exploring graph encoders with attention for encoding AMRs \cite{nguyenetal2021}. In this study, various neural machine translation models were examined, including seq-to-seq models with Bi-LSTM and transformers, with the latter reported as less effective than the former in this context.

AMR-NMT \cite{li-flanigan-2022-improving} integrates AMRs into the transformer architecture for neural machine translation. The proposed model combines a transformer with a heterogeneous graph transformer, enabling it to encode both source sentences and their corresponding AMR graphs. The architecture (AMR-Transformer) features parallel stacked encoder and decoder layers: one set processes the sequence data, while the other handles the graph data. The two resulting representations are then integrated to generate the final translation. This approach leverages the strengths of both sequence encoding and graph-based semantic encoding, resulting in improved translation quality. Experimental results indicate that the AMR-Transformer outperforms both the standard transformer model and previous non-transformer-based models across two different language pairs in high-resource and low-resource settings.

A novel application of AMRs in machine translation was explored by Wein et al. \cite{wein-schneider-2024-lost} in the context of translationese reduction. The term ``translationese'' refers to linguistic features, such as semantic patterns and unique syntactic structures, that are characteristic of human-translated texts. Reducing translationese can enhance the evaluation of machine translation models. The proposed parse-then-generate technique first parses a text affected by translationese into an AMR and then generates text from that AMR. Although the analysis revealed that AMR generation struggles to produce fluent texts, it remains the only method (compared to T5-based and BART-based baselines) that contributes to translationese reduction, thereby confirming the promise of AMRs as an Interlingua.

\subsection{Text Classification}
\begin{figure}[t]
    \centering
    \includegraphics[width=\textwidth]{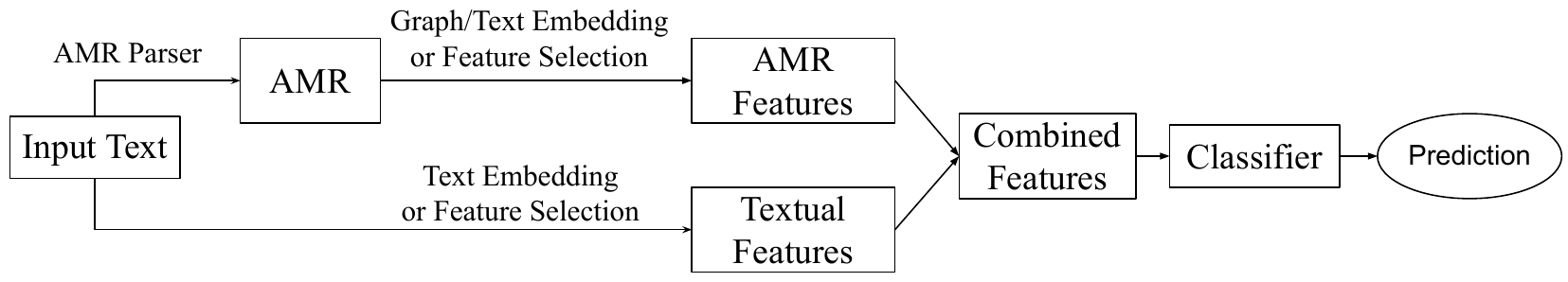}
    \caption{Application of AMR for text classification. The input text is parsed to generate the AMR. Features from AMR can be derived from linearization and text embeddings, graph embeddings, or other designed features. The textual features are extracted from raw text and then combined with AMR features. These features are passed to a classifier to predict a label.}
    \label{fig:ClassificationAMR}
\end{figure}

Text classification involves assigning categories to textual data, supporting tasks such as sentiment analysis, paraphrase detection, and fake news detection. The proposed models with AMRs often combine the AMR features with textual features for classification, as shown in Fig. \ref{fig:ClassificationAMR}. For instance, FakEDAMR \cite{10.1007/978-3-031-53468-3_26} encodes textual content using AMRs for fake news detection. Using GloVe embeddings for text, and graph embeddings of AMRs, these embeddings are concatenated and passed to a BiLSTM classification layer. Another example is AMR-CNN \cite{10243627} which integrates AMR with a Convolutional Neural Network (CNN) to detect the level of profanity and toxicity in online text content. Unlike traditional toxic content detection systems that rely on lexicon- and keyword-based methods, AMR-CNN focuses on understanding the underlying meaning of sentences. AMR allows for the extraction of semantic roles, entity types, coreference resolution, and other nuanced aspects of language, such as modality and polarity, making it well-suited for the task of detecting abusive or toxic content.

For sentiment analysis, the AMR-based Path Aggregation Relational Network (APARN) \cite{ma-etal-2023-amr} model investigates replacing dependency trees with AMRs. In this model, the input text is first parsed into AMR with the SPRING model \cite{Bevilacqua_Blloshmi_Navigli_2021}. Then, the AMR is aligned using LEAMR \cite{blodgett-schneider-2021-probabilistic} to rebuild AMR relations between words in the sentence and AMR nodes. Using BERT, embeddings for words in sentences, as well as AMR nodes and edges are generated. These embeddings are passed to the path aggregator component that combines them into a relational feature matrix. Finally, the output weight matrix from the path aggregator is added to the self-attention matrix from the original sentence and passed to a fully connected softmax layer to classify sentiment. The proposed approach achieved a higher accuracy compared to using BERT alone, with improvements of 2-5\% across different datasets).

AMRs also serve as a suitable representation for Natural Language Inference (NLI), the task of determining whether a given premise entails, contradicts, or is neutral with respect to a hypothesis.
AMR4NLI \cite{opitz-etal-2023-amr4nli} studied whether AMRs could represent the premises and hypotheses more effectively (through WWLK metric), allowing a more interpretable way to measure if the hypothesis is a semantic substructure of the premise. The results show that while AMRs capture the semantic structure and relationships explicitly, contextualized embeddings (through BERTScore) provide a rich, dense representation of meaning. By combining these two through a linear combination of graph and text similarity metrics, they created a hybrid model that leveraged the strengths of both approaches. This hybrid model was tested across several English NLI benchmarks, showing improved robustness. 

Another task where AMRs are beneficial is paraphrase detection, deciding whether two sentences are paraphrases of each other. Issa et al. \cite{issa-etal-2018-abstract} explored AMRs for this task. While the naïve application of AMR parsing (as a bag of words with SMatch) did not yield satisfactory results, the authors combined AMR with Latent Semantic Analysis (LSA) to better capture the semantic similarities between sentences. LSA creates a sentence-term matrix based on TF-IDF, and AMR is used to re-weight the values in this matrix. By applying PageRank to the AMR graphs, the most influential nodes were identified in each graph. This helped in focusing on the most critical parts of the sentence when determining whether two sentences were paraphrases. 

To build paraphrase dataset, the PARAAMR \cite{huang-etal-2023-paraamr} leverages AMR-to-text generation to generate syntactically diverse paraphrases. Traditional methods of generating paraphrases, such as machine back-translation, often lack syntactic variety, generating sentences that are too similar structurally. PARAAMR addresses this by using AMR to create paraphrases with greater syntactic diversity while maintaining semantic consistency. In this approach, the AMR for input text is modified by changing the focus. To change the focus, nodes are randomly selected as the new focus, and the incoming edges for that node are reversed. The resulting AMR is then passed to an AMR-to-text generation pipeline to produce the paraphrase. Another approach to generate paraphrases with AMR is SAPG (``Semantically-Aware Paraphrase Generation with AMR Graphs'') \cite{sousa-cardoso-2025-sapg} that encodes AMR graphs of input texts using a graph neural network-based encoder integrated with a pretrained language model. SAPG employs a dual-encoder architecture. One encoder processes a linearized version of the AMR graph, while the other uses a Graph Neural Network (GNN) to encode the AMR graph structure directly into a pretrained language model. 

A recent similar application to paraphrase detection was used in DART model \cite{park-etal-2025-dart} to detect AI-generated text. The core idea is that rephrasing a text can reveal semantic differences between human and AI writing styles. DART leverages AMR as a semantic representation to capture the underlying meaning of both the original text and its rephrased versions. The framework parses texts into AMR graphs and then uses SEMA (recall and precision) between the AMR of the original text and the AMRs of its rephrased counterparts. These similarity scores are then passed to a classifier to detect AI-generated text. The experimental results shows this model achieve the highest F1-score compared to state-of-the-art models.

\subsection{Information Extraction}
Information Extraction (IE) aims to extract structured information from unstructured text, but traditional methods often struggle to capture deep semantics. AMR enhances IE by providing semantic graphs that capture concepts, relationships, and roles, enabling more accurate event and argument extraction across various domains. For example, Li et al. \cite{li-etal-2015-improving-event} explored event detection with AMR to identify and classify events. The AMR graph is analyzed to identify event triggers, which are words or phrases that indicate the occurrence of an event. To this end, a Maximum Entropy classifier is used for event classification using textual features (e.g., lexical features ), along with AMR features. The AMR features are defined around the trigger node (which indicates the occurrence of an event) and include features such as the distance to the root, the node value, the parent node, sibling nodes, and children. The analysis shows that classification accuracy can improve by 2\% when AMR features are incorporated.  

The TSAR (Two-Stream Abstract meaning Representation) \cite{xu-etal-2022-two} model presents a novel approach to improve the extraction of event arguments from documents. Event argument extraction focuses on identifying the entities that serve as event arguments and predicting their roles in events. The two-stream encoding module in this work employs both global and local encoders to capture different levels of context. First, with the same BERT-based pre-trained model, global and local embeddings are generated from the input text. Two types of AMRs are then generated; one as a local AMR based on sentences, and the other as a global AMR by connecting the roots of local AMRs. These AMRs are used by the AMR-guided interaction module to generate local and global embeddings, which are then fused and passed to a classifier. 

% global graph on by combining t
% AMR graphs and local ones to stimulate the interactions among concepts in the document, especially those far
% away from each other, based on graph neural network. Next, the information fusion module fuses the two-stream
% representations, and also strengthens the boundary information through a boundary loss. Finally, the classification
% module makes predictions for candidate spans. 

% These encoders share the same BERT-based pre-trained model, and their generated encodings serve as the initial node embeddings for the AMR graph. The AMR-guided interaction module has three components. First, the embedding for a node in AMR is composed by averaging the local representations from corresponding text generated from the previous step. Then an L-layer stacked Graph Convolution Network \cite{kipf2017semisupervised} is used to generate embeddings for concept nodes in the AMRs to capture their interactions. Finally, the embeddings are decomposed back to the text span. The same components are used for the whole document, by attaching the roots of all the sentences AMRs and using the global encoder for initialization. The global and local features of a document are then fused and used for classification.

AMRs are also used for the identification of biomedical events, defined as the automatic identification and classification of specific biological events, such as gene expressions and regulatory relationships \cite{kim-etal-2011-overview-genia}. Rao et al. viewed this as a subgraph identification problem \cite{rao-etal-2017-biomedical}. They developed a neural network model to identify these event subgraphs and employed a distant supervision technique to generate additional training data. Their approach was evaluated using the 2013 Genia Event Extraction dataset, yielding promising results for the application of AMR in biomedical text analysis. The Knowledge-AMR \cite{zhang-etal-2021-fine} is another model that enriches AMRs with an external knowledge base to make them more effective for extracting fine-grained biomedical information. This involves incorporating biomedical knowledge into AMR to improve its expressiveness and accuracy in representing the semantics of biomedical texts. The results show that the enriched AMR can be effective for extracting complex biomedical entities and relations, including fine-grained relations between biological processes, molecular functions, and disease mechanisms from literature, which is typically more challenging in biomedical texts.

Another similar application is drug-drug interactions (DDI). DDI studies evaluate how the presence of one drug alters the effects of another. Wang et al. \cite{10.1145/3107411.3107426} viewed DDI as a binary classification task, with drug-drug pairs are classified as either positive (indicating interaction) or negative. Since many drug names may not appear in the AMR bank, first the word ``drug'' is replaced with ``medication'', and then all drug entities are substituted with the word ``drug''. DDI sentences are then parsed with JAMR into AMR graphs, followed by restoring the drug entities. After that, the AMR is passed to the SkipGram model to generate AMR embeddings. Along with AMR embeddings, dependency parse graphs, and raw text embedding features were used for classification. The results indicate that while AMRs can provide useful information, they should be used with raw text embeddings for improvement to yield better effectiveness.

\subsection{Information Seeking}
%Information Retrieval and Question Answering

AMR is effective for information retrieval and question answering because it captures the underlying semantic structure of sentences, enabling a deeper understanding of meaning. By representing concepts, relationships, and events in a graph-like structure, AMR allows for more accurate matching of user queries with relevant information, especially in complex domains like biomedicine and mathematics. For example, InfoForager \cite{bonial-etal-2020-infoforager} uses AMR to find answers related to COVID-19. Sentences in the collection are first parsed into their AMR graphs, which are then matched against the input question AMR using SMatch for ranking retrieved sentences. 

MathAMR \cite{10.1145/3511808.3557567} integrates mathematical formula representations with AMRs for the contextualized formula search, forming a unified search representation (as shown in Fig. \ref{fig:mathAMR}). The goal of this task is to find relevant formulae to a formula query within the same context. Math formulas are usually represented in the form of operator trees, which show what operations are applied to what operands. This representation is very similar to AMRs for raw text. Therefore, in this approach, first formulas are masked with placeholders and their context is parsed to an AMR. The operator tree is then reconnected to the AMR by linking the root of the operator tree to the placeholder node for the formula. For ranking, the AMRs of the query and candidates are linearized with the depth-first search (ignoring edge labels), and the Sentence-BERT \cite{reimers-gurevych-2019-sentence} model is used to calculate the similarity.

AMRs have also been explored for semantic information retrieval in the domain of remote sensing and image exploitation \cite{9931702}. The authors adapted and evaluated an AMR-based search approach to find relevant documents and extract pertinent content based on natural language queries. SMatch and SemBleu were studied as AMR similarity metrics to determine their suitability for this task. The study found that while both metrics yield promising results in ranking sentences, SemBleu is more viable for practical applications due to its lower sensitivity to grammatical variations and its ability to handle synonyms effectively.
% This work highlights the potential of using AMR for semantic search in large data management systems and provides insights into the strengths and weaknesses of SMatch and SemBleu as comparison metrics.

\begin{figure}[t]
    \centering
    \includegraphics[width=\textwidth]{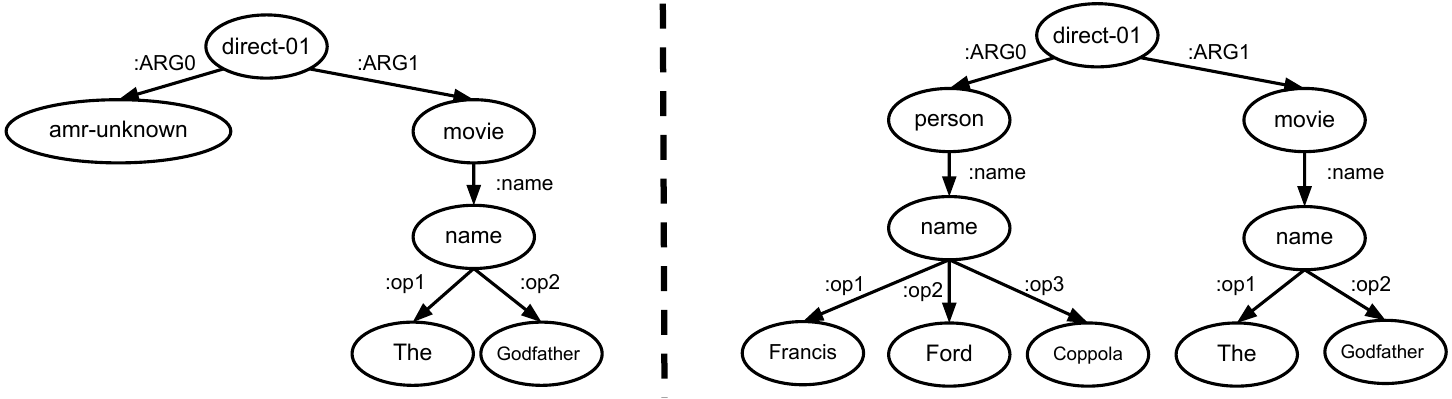}
    \caption{Application of AMR for Question-Answering. The AMR on the left corresponds to the question `Who directed The Godfather movie?' and the AMR on the right represents the answer `The Godfather is directed by Francis Ford Coppola.'}
    \label{fig:QAAMR}
\end{figure}

AMRs are further explored for question answering. A special node, `amr-unknown,' is used in place of the answer to the question, making this representation particularly suitable for this tasks. For instance, for the question `Who directed The Godfather movie?'  and its corresponding answer `The Godfather is directed by Francis Ford Coppola.' the AMRs are shown in Fig. \ref{fig:QAAMR}. As shown, the `amr-unknown' node in the question AMR can be matched with the left subgraph from the answer AMR, while the root and the right subgraphs remain the same.

Building on these capabilities, Neuro-Symbolic Question Answering (NSQA) \cite{kapanipathi-etal-2021-leveraging} uses AMRs to improve question understanding and logical query formation in Knowledge Base Question Answering (KBQA). KBQA is the task of answering natural language questions based on facts in a knowledge base. NSQA generates AMR for a given question, and then uses a novel graph transformation technique to generate candidate logical queries aligned with a knowledge base. These logical queries are then Logical Neural Network to produce the final answer. Another example is the Question Decomposition method based on AMR (QDAMR) \cite{ijcai2022p0568}, introduced for multi-hop question answering. QDAMR utilizes AMR to achieve interpretable reasoning by decomposing complex, multi-hop questions into simpler sub-questions. This is achieved by segmenting the AMR graph based on reasoning types. The resulting sub-questions are answered sequentially using an existing QA model. This approach not only improves multi-hop QA performance but also generates well-formed sub-questions that outperform those produced by other question decomposition methods. By integrating interpretability directly into the QA process, QDAMR provides a clearer reasoning path. However, future research is required to assess the utility of its explanations for human users.

% The study highlights NSQA’s modular architecture as a key advantage, allowing for task-specific training and adaptability to future extensions, such as temporal reasoning and handling KB incompleteness.

\subsection{Other Applications}
While we have explored some of the common downstream applications of AMRs, their utility extends to many other areas. These include tasks such as fact verification \cite{jayaweera-etal-2024-amrex, qiu-etal-2024-amrfact}, legal judgement prediction \cite{vijay-hershcovich-2024-abstract}, and time-line generation \cite{10.1145/3543873.3587670}. AMRs have also been applied to data augmentation. AMR-DA \cite{shou-etal-2022-amr} leverages AMRs to enhance training datasets by converting sentences into AMR graphs, applying augmentation strategies to modify these graphs, and generating new sentence variations from the altered graphs. AMR-DA integrates sentence-level techniques like back translation with token-level approaches such as Easy Data Augmentation (EDA) to create a more robust data augmentation framework. The method is evaluated on two tasks: semantic textual similarity (STS) and text classification. Results show that AMR-DA enhances state-of-the-art models' performance on multiple STS benchmarks and outperforms existing augmentation techniques like EDA in text classification tasks. This work is the first to utilize AMR for data augmentation, showing that it can effectively diversify training data and improve model robustness without requiring decoder retraining.

Another generative data augmentation methodology, ABEX \cite{ghosh-etal-2024-abex}, focuses on enhancing low-resource Natural Language Understanding (NLU) tasks. ABEX employs a two-step process: first, it converts documents into AMR graphs to capture their semantic structure, and then it edits these graphs to generate abstract descriptions by removing non-essential details while retaining the core meaning. This approach is training-free and flexible, enabling it to adapt to various tasks without the need for extensive model retraining. ABEX also addresses data scarcity by synthesizing large-scale datasets of abstract-document pairs, using prompts to large language models (LLMs) to create abstract descriptions. Its innovative use of AMR editing and generative augmentation enhances data diversity and quality, providing a robust solution for low-resource scenarios.

% The methodology's novelty lies in its combination of AMR editing and generative data augmentation, which not only improves the diversity and quality of generated outputs but also closely mimics human language processing, setting it apart from traditional augmentation techniques.

AMRs are also used for image-text processing, particularly for tasks like dense image captioning. Dense captioning involves generating multiple textual descriptions for various regions of an image. For instance, an image of a girl riding a bicycle could yield descriptions such as ``a girl riding a bicycle,'' ``the girl who rides a bicycle,'' or ``the girl bicycle rider.'' To address the variability in descriptions, Antonio et al. \cite{10.1007/978-3-030-61377-8_31} utilizes DenseCap \cite{Johnson_2016_CVPR}, comparing training the model with natural language and different AMR variants: (1) linearized AMR, (2) anonymized linearized AMR (replacing named entities and quantifiers with placeholders), and (3) concatenated anonymized AMR (merging concepts and arguments). For evaluation, natural language targets were converted to AMRs using the same pipeline. Results showed that the concatenated anonymized AMR input was most effective, achieving nearly double the mean average precision (mAP) compared to natural language. Future work could investigate using AMR-to-text models for a fair comparison with natural text.

For visual scene understanding, Meta-AMR graphs \cite{abdelsalam-etal-2022-visual} address the limitations of traditional scene graphs (SGs). Scene graphs abstract visual inputs into structured representations, focusing on spatial relationships between entities, but they require extensive manual annotation and lack higher-level semantic information. In contrast, AMR graphs capture abstract semantic concepts, offering a richer understanding of scenes. The proposed approach parses images into AMRs by leveraging text-to-AMR parsers and introduces meta-AMR graphs to unify information from multiple image descriptions. The methodology involves two models: one that directly generates AMRs from images and another that operates in two stages, predicting nodes first and then their relationships. Both models use visual features extracted from images and are trained on datasets pairing images with captions and corresponding AMR graphs, enhancing the representation and understanding of visual scenes.

\section{Conclusion: Then, Now, Future}
This survey explored Abstract Meaning Representation (AMR), its intended use, and its applications in natural language processing tasks. AMR serves as a semantic representation that captures ``who is doing what to whom.'' AMR parsers generate these representations for a given text, with current state-of-the-art models predominantly based on seq-to-se architectures. The application of large language models (LLMs) for AMR parsing is an emerging area of research. While there have been efforts to develop parsers capable of handling longer text spans, most AMR parsers remain focused on sentence-level parsing. Furthermore, although AMR was not initially designed for non-English languages, several approaches have been proposed to extend its capabilities to multilingual parsing.

AMRs have continued to evolve over time, with various extensions and enrichment strategies being introduced. One notable extension is Uniform Meaning Representation (UMR) \cite{VanGysel2021}, which aims to support cross-linguistic plausibility and portability. UMRs are currently a focus of active research, and it is possible that they could eventually replace traditional AMRs in some applications.

Text generation from AMRs is another domain of active research discussed in this survey. This task focuses on generating corresponding text from an AMR while preserving the semantic integrity of the original input. Similar to AMR parsing, current approaches for text generation are largely based on sequence-to-sequence models, with LLMs being increasingly explored for this purpose. As AMRs become more integrated into text generation workflows, they may eventually be incorporated into LLMs to improve the semantic quality of generated text.

Finally, we explored various downstream applications of AMRs, including text-to-text generation, text classification, information extraction, information seeking, and tasks like caption generation. Initially designed to abstract meaning, AMRs were first applied to related tasks such as summarization. However, they are now gaining traction in multimodal tasks, where AMR features are combined with data types such as images. Another emerging application is data augmentation, where pipelines generate text from AMRs to enrich training datasets.

Looking ahead, AMRs are likely to find novel applications in tasks such as fairness, visual question-answering, and story generation. By abstracting and representing meaning effectively, AMRs have the potential to provide valuable features that enhance system performance and improve task outcomes.

% We have seen where AMRs started and how the related research is going. As AMRs evolve through time, with advances in artificial intelligence models, 
% AMR for LLMs \cite{yang2024emphasisingstructuredinformationintegrating}

% For a more comprehensive list of AMR papers, we refer our reader to \url{https://nert-nlp.github.io/AMR-Bibliography/}.

% \cite{lai-etal-2020-continuation}
% Low-Resource Generation \cite{sobrevilla-cabezudo-etal-2024-investigating-paraphrase}

% Despite advances, for AMR-to-text generation task, the Blue score remains around 0.5 with METEOR close to 45.0.

% \begin{verbatim}
%   \appendix
% \end{verbatim}
% \begin{acks}
% \end{acks}

\bibliographystyle{ACM-Reference-Format}
\bibliography{sample-base}
\clearpage
\appendix

\section{Resources and Tools}
Table \ref{tab:resources} provides a list of AMR resources and tools.
\begin{table}[b]
  \centering
  \caption{Selected Papers and Resources for Abstract Meaning Representation.}
  \label{tab:resources}
  \resizebox{0.9\textwidth}{!}{
  \begin{tabular}{l|l|l|c}
    \toprule
       Model&Application & Link&Year\\
     \hline
    \multicolumn{3}{l}{Parsers and Generators}\\
    \hline
    JAMR \cite{flanigan-etal-2014-discriminative}& Parser, Generator &\url{https://github.com/jflanigan/jamr}&2014\\
    CAMR \cite{wang-etal-2016-camr}& Parser, Generator &\url{https://github.com/c-amr/camr}&2016\\
    AMR-Eager \cite{damonte-etal-2017-incremental}& Parser-Multilingual, Generator &\url{https://github.com/mdtux89/amr-eager}&2017\\
    NeuralAMR \cite{konstas-etal-2017-neural} &Parser, Generator&\url{https://github.com/sinantie/NeuralAmr}&2017\\
    AMR-GP \cite{lyu-titov-2018-amr}& Parser& \small\url{https://github.com/ChunchuanLv/AMR_AS_GRAPH_PREDICTION}&2018\\
    Graph-to-sequence \cite{song-etal-2018-graph} & Generator&\url{https://github.com/freesunshine0316/neural-graph-to-seq-mp}&2018\\
     Structural Transformer \cite{zhu-etal-2019-modeling} & Generator&\url{https://github.com/Amazing-J/structural-transformer}&2019\\
    STOG \cite{zhang-etal-2019-amr}& Parser &\url{https://github.com/sheng-z/stog}&2019\\
   GraphTransformer \cite{wang-etal-2020-amr}& Generator&\url{https://github.com/sodawater/GraphTransformer}&2020\\
    Graph Reconstruction \cite{ijcai2020p542}& Generator& \small\url{https://github.com/sodawater/graph-reconstruction-amr2text}&2020\\
    GPT-too \cite{mager-etal-2020-gpt}& Generator& \url{https://github.com/IBM/GPT-too-AMR2text}&2020\\
    XLPT-AMR \cite{xu-etal-2021-xlpt} & Parser, Generator-Multilingual & \url{https://github.com/xdqkid/XLPT-AMR} & 2021\\
    XAMR \cite{cai-etal-2021-multilingual-amr} & Parser-Multilingual&\url{https://github.com/jcyk/XAMR} &2021\\
    APT \cite{zhou-etal-2021-amr} & Parser&\url{https://github.com/IBM/transition-amr-parser}&2021\\
    SPRING \cite{bevilacqua2021one} &Parser, Generator&\url{https://github.com/SapienzaNLP/spring}&2021\\
    StructAdapt \cite{ribeiro-etal-2021-structural}&Generator&\url{https://github.com/UKPLab/StructAdapt}&2021\\
    AMRBART \cite{bai-etal-2022-graph}&Parser, Generator&\url{https://github.com/goodbai-nlp/AMRBART}&2022\\
     BiBL\cite{cheng-etal-2022-bibl}&Parser, Generator&\url{https://github.com/KHAKhazeus/BiBL}&2022\\
     XL-AMR \cite{blloshmi-etal-2020-xl} & Parser-Multilingual&\url{https://github.com/SapienzaNLP/xl-amr} &2022\\
     EnrichedAMR \cite{ji-etal-2022-automatic} & Enrichment & \url{https://github.com/emorynlp/EnrichedAMR/}& 2022\\
     LeaKDistill \cite{vasylenko-etal-2023-incorporating}& Parsing & \url{https://github.com/sapienzanlp/LeakDistill}&2023\\
     Meta-XMAR \cite{kang-etal-2024-cross-lingual}& Parser-Multilingual&\url{https://github.com/Emvista/Meta-XAMR-2024} &2024\\
    \hline
    \multicolumn{3}{l}{Evaluation Tools}\\
    \hline
    SMatch \cite{cai-knight-2013-SMatch} & Evaluation & \url{https://github.com/snowblink14/SMatch} & 2013\\
    SemBlue \cite{song-gildea-2019-sembleu} & Evaluation & \url{https://github.com/freesunshine0316/sembleu}&2019\\
    SEMA \cite{anchieta2019semaextendedsemanticevaluation} & Evaluation& \url{https://github.com/rafaelanchieta/sema}&2019\\
    S$^2$match \cite{opitz-etal-2020-amr}& Evaluation & \url{https://github.com/Heidelberg-NLP/amr-metric-suite} & 2020\\
    (WWLK) \cite{opitz-etal-2021-weisfeiler} & Evaluation & \url{https://github.com/Heidelberg-NLP/weisfeiler-leman-bamboo}&2021\\
    XS$^2$match \cite{feng-etal-2022-language}&Evaluation-Multilingual&\url{https://github.com/shirawein/Crossling-AMR-Eval} & 2022\\
    SMatch++ \cite{opitz-2023-SMatch} & Evaluation & \url{https://github.com/flipz357/SMatchpp}&2023\\
    AMRSim \cite{shou-lin-2023-evaluate}& Evaluation & \url{https://github.com/zzshou/AMRSim}&2023\\
    GrAPES \cite{groschwitz-etal-2023-amr} & Evaluation & \url{https://github.com/jgroschwitz/GrAPES} & 2023\\
    \hline
    \multicolumn{3}{l}{Corpus}\\
    \hline
    AMR 1.0 \cite{LDC1}& Corpus&\url{https://catalog.ldc.upenn.edu/LDC2014T12}&2014\\
    AMR 2.0 \cite{LDC2}& Corpus&\url{https://catalog.ldc.upenn.edu/LDC2017T10}&2017\\
    AMR 3.0 \cite{LDC3}& Corpus&\url{https://catalog.ldc.upenn.edu/LDC2020T02}&2020\\
    CHAMR \cite{li-etal-2016-annotating} &Corpus-Chinese&\url{https://www.cs.brandeis.edu/~clp/camr/camr.html}& 2016\\
    % QALD-9-AMR \cite{lee-etal-2022-maximum} & Corpus &\url{https://github.com/IBM/AMR-annotationsSMatch} & 2022\\
    AMR 2.0 - 4 translation \cite{AB2/5OU0AQ_2022}&Corpus-Multilingual&\url{https://hdl.handle.net/11272.1/AB2/5OU0AQ}&2022\\
   MASSIVE-AMR \cite{regan2024massivemultilingualabstractmeaning}&Corpus-Multilingual&\url{https://github.com/amazon-science/MASSIVE-AMR}&2024\\
    DeAMR \cite{otto-etal-2024-corpus} & Corpus-German & \url{https://github.com/chriott/DeAMR/}&2024\\
    \hline
    \multicolumn{3}{l}{Applications}\\
    \hline
    SemanticSummarizer \cite{liu-etal-2015-toward} & Summarization & \url{https://github.com/summarization/semantic_summ}& 2015\\
    AMR2Text-Summarizer \cite{hardy-vlachos-2018-guided} & Summarization & \url{https://github.com/sheffieldnlp/AMR2Text-summ} &2018\\
    TSAR \cite{xu-etal-2022-two}&Event Detection&\url{https://github.com/PKUnlp-icler/TSAR}&2022\\
    Knowledge-AMR \cite{zhang-etal-2021-fine}&Information Extraction & \url{https://github.com/zhangzx-uiuc/Knowledge-AMR} & 2021\\
    Densecap-amr  \cite{10.1007/978-3-030-61377-8_31}&Dense Captioning &\url{https://github.com/LALIC-UFSCar/densecap-amr}&2020\\
    APARN \cite{ma-etal-2023-amr}& Sentiment Analysis &\url{https://github.com/THU-BPM/APARN}&2023\\
    AMR4NLI \cite{opitz-etal-2023-amr4nli}&Natural Language Inference&\url{https://github.com/flipz357/AMR4NLI}&2023\\
 AMR-DA \cite{shou-etal-2022-amr}&Data Augmentation&\url{https://github.com/zzshou/amr-data-augmentation}&2022\\
    AMR-NMT \cite {li-flanigan-2022-improving} & Machine Translation & \url{https://github.com/jlab-nlp/amr-nmt}&2022\\
AMR-Translationese \cite{wein-schneider-2024-lost} & Machine Translation & \url{https://github.com/shirawein/amr-translationese}&2024\\
ABEX \cite{ghosh-etal-2024-abex}&Data Augmentation&\url{https://github.com/Sreyan88/ABEX}&2024\\
SAPG \cite{sousa-cardoso-2025-sapg}&Paraphrase Generation & \url{https://github.com/afonso-sousa/sapg}&2025\\
    \bottomrule
  \end{tabular}
  }
\end{table}
\end{document}